\documentclass[10pt,twocolumn,letterpaper]{article}

\usepackage[pagenumbers]{cvpr}
\usepackage[accsupp]{axessibility}  % Improves PDF readability for those with disabilities.

\makeatletter
\@namedef{ver@everyshi.sty}{}
\makeatother
\usepackage{tikz}
\usetikzlibrary{spy,calc}

\usepackage{graphicx}
\usepackage{amsmath}
\usepackage{amssymb}

\usepackage{booktabs}
\usepackage{multirow} % multirow cells in tabular
\usepackage[flushleft]{threeparttable}
\usepackage{cuted}  % Mix 1-col, 2-col content
\usepackage{capt-of}
\usepackage[export]{adjustbox}  % For \adjincludegraphics (cropping)
\usepackage[percent]{overpic}  % Text overlay onto picture

\usepackage{xcolor}
\usepackage{bm}
\usepackage{pifont}  % http://ctan.org/pkg/pifont
\usepackage{mathtools}

\usepackage[pagebackref,breaklinks,colorlinks]{hyperref}
\usepackage[capitalize]{cleveref}  % \cref commands equals \Cref

\usepackage{siunitx}
\robustify\bfseries
\newif\ifblackandwhitecycle
\gdef\patternnumber{0}

\pgfkeys{/tikz/.cd,
    zoombox paths/.style={
        draw=orange,
        very thick
    },
    black and white/.is choice,
    black and white/.default=static,
    black and white/static/.style={
        draw=white,
        zoombox paths/.append style={
            draw=white,
            postaction={
                draw=black,
                loosely dashed
            }
        }
    },
    black and white/static/.code={
        \gdef\patternnumber{1}
    },
    black and white/cycle/.code={
        \blackandwhitecycletrue
        \gdef\patternnumber{1}
    },
    black and white pattern/.is choice,
    black and white pattern/0/.style={},
    black and white pattern/1/.style={
            draw=white,
            postaction={
                draw=black,
                dash pattern=on 2pt off 2pt
            }
    },
    black and white pattern/2/.style={
            draw=white,
            postaction={
                draw=black,
                dash pattern=on 4pt off 4pt
            }
    },
    black and white pattern/3/.style={
            draw=white,
            postaction={
                draw=black,
                dash pattern=on 4pt off 4pt on 1pt off 4pt
            }
    },
    black and white pattern/4/.style={
            draw=white,
            postaction={
                draw=black,
                dash pattern=on 4pt off 2pt on 2 pt off 2pt on 2 pt off 2pt
            }
    },
    zoomboxarray inner gap/.initial=5pt,
    zoomboxarray columns/.initial=2,
    zoomboxarray rows/.initial=1,
    zoomboxarray heightmultiplier/.initial=0.5,
    subfigurename/.initial={},
    figurename/.initial={zoombox},
    zoomboxarray/.style={
        execute at begin picture={
            \begin{scope}[
                spy using outlines={%
                    zoombox paths,
                    width=\imagewidth / \pgfkeysvalueof{/tikz/zoomboxarray columns} - (\pgfkeysvalueof{/tikz/zoomboxarray columns} - 1) / \pgfkeysvalueof{/tikz/zoomboxarray columns} * \pgfkeysvalueof{/tikz/zoomboxarray inner gap} -\pgflinewidth,
                    height=\pgfkeysvalueof{/tikz/zoomboxarray heightmultiplier} * (\imageheight / \pgfkeysvalueof{/tikz/zoomboxarray rows} - (\pgfkeysvalueof{/tikz/zoomboxarray rows} - 1) / \pgfkeysvalueof{/tikz/zoomboxarray rows} * \pgfkeysvalueof{/tikz/zoomboxarray inner gap}-\pgflinewidth),
                    magnification=3,
                    every spy on node/.style={
                        zoombox paths
                    },
                    every spy in node/.style={
                        zoombox paths
                    }
                }
            ]
        },
        execute at end picture={
            \end{scope}
     \gdef\patternnumber{0}
        },
        spymargin/.initial=0.5em,
        zoomboxes xshift/.initial=1,
        zoomboxes right/.code=\pgfkeys{/tikz/zoomboxes xshift=1},
        zoomboxes left/.code=\pgfkeys{/tikz/zoomboxes xshift=-1},
        zoomboxes yshift/.initial=0,
        zoomboxes above/.code={
            \pgfkeys{/tikz/zoomboxes yshift=1},
            \pgfkeys{/tikz/zoomboxes xshift=0}
        },
        zoomboxes below/.code={
            \pgfkeys{/tikz/zoomboxes yshift=-1},
            \pgfkeys{/tikz/zoomboxes xshift=0}
        },
        caption margin/.initial=0ex, %
    },
    adjust caption spacing/.code={},
    image container/.style={
        inner sep=0pt,
        at=(image.north),
        anchor=north,
        adjust caption spacing
    },
    zoomboxes container/.style={
        inner sep=0pt,
        at=(image.north),
        anchor=north,
        name=zoomboxes container,
        xshift=\pgfkeysvalueof{/tikz/zoomboxes xshift}*(\imagewidth+\pgfkeysvalueof{/tikz/spymargin}),
        yshift=\pgfkeysvalueof{/tikz/zoomboxes yshift}*(\imageheight+\pgfkeysvalueof{/tikz/spymargin}+\pgfkeysvalueof{/tikz/caption margin}),
        adjust caption spacing
    },
    calculate dimensions/.code={
        \pgfpointdiff{\pgfpointanchor{image}{south west} }{\pgfpointanchor{image}{north east} }
        \pgfgetlastxy{\imagewidth}{\imageheight}
        \global\let\imagewidth=\imagewidth
        \global\let\imageheight=\imageheight
        \gdef\columncount{1}
        \gdef\rowcount{1}
        
    },
    image node/.style={
        inner sep=0pt,
        name=image,
        anchor=south west,
        append after command={
            [calculate dimensions]
            node [image container,subfigurename=\pgfkeysvalueof{/tikz/figurename}-image] {\phantomimage}
            node [zoomboxes container,subfigurename=\pgfkeysvalueof{/tikz/figurename}-zoom] {\phantomimage}
        }
    },
    color code/.style={
        zoombox paths/.append style={draw=#1}
    },
    connect zoomboxes/.style={
    spy connection path={\draw[draw=none,zoombox paths] (tikzspyonnode) -- (tikzspyinnode);}
    },
    help grid code/.code={
        \begin{scope}[
                x={(image.south east)},
                y={(image.north west)},
                font=\footnotesize,
                help lines,
                overlay
            ]
            \foreach \x in {0,1,...,9} {
                \draw(\x/10,0) -- (\x/10,1);
                \node [anchor=north] at (\x/10,0) {0.\x};
            }
            \foreach \y in {0,1,...,9} {
                \draw(0,\y/10) -- (1,\y/10);                        \node [anchor=east] at (0,\y/10) {0.\y};
            }
        \end{scope}
    },
    help grid/.style={
        append after command={
            [help grid code]
        }
    },
}

\newcommand\phantomimage{%
    \phantom{%
        \rule{\imagewidth}{\imageheight}%
    }%
}
\newcommand\zoombox[2][]{
    \begin{scope}[zoombox paths]
        \pgfmathsetmacro\xpos{
            (\columncount-1)*(\imagewidth / \pgfkeysvalueof{/tikz/zoomboxarray columns} + \pgfkeysvalueof{/tikz/zoomboxarray inner gap} / \pgfkeysvalueof{/tikz/zoomboxarray columns} ) + \pgflinewidth
        }
        \pgfmathsetmacro\ypos{
            (\rowcount-1) * (\imageheight / \pgfkeysvalueof{/tikz/zoomboxarray rows} + \pgfkeysvalueof{/tikz/zoomboxarray inner gap} / \pgfkeysvalueof{/tikz/zoomboxarray rows} ) + 0.5*\pgflinewidth
        }
        \edef\dospy{\noexpand\spy [
            #1,
            zoombox paths/.append style={
                black and white pattern=\patternnumber
            },
            every spy on node/.append style={#1},
            x=\imagewidth,
            y=\imageheight
        ] on (#2) in node [anchor=north west] at ($(zoomboxes container.north west)+(\xpos pt,-\ypos pt)$);}
        \dospy
        \pgfmathtruncatemacro\pgfmathresult{ifthenelse(\columncount==\pgfkeysvalueof{/tikz/zoomboxarray columns},\rowcount+1,\rowcount)}
        \global\let\rowcount=\pgfmathresult
        \pgfmathtruncatemacro\pgfmathresult{ifthenelse(\columncount==\pgfkeysvalueof{/tikz/zoomboxarray columns},1,\columncount+1)}
        \global\let\columncount=\pgfmathresult
        \ifblackandwhitecycle
            \pgfmathtruncatemacro{\newpatternnumber}{\patternnumber+1}
            \global\edef\patternnumber{\newpatternnumber}
        \fi
    \end{scope}
}

\definecolor{turquoise}{cmyk}{0.65,0,0.1,0.3}
\definecolor{purple}{rgb}{0.65,0,0.65}
\definecolor{dark_green}{rgb}{0, 0.5, 0}
\definecolor{orange}{rgb}{0.8, 0.6, 0.2}
\definecolor{darkred}{rgb}{0.6, 0.1, 0.05}
\definecolor{blueish}{rgb}{0.0, 0.3, .6}
\definecolor{light_gray}{rgb}{0.7, 0.7, .7}
\definecolor{pink}{rgb}{1, 0, 1}
\definecolor{greyblue}{rgb}{0.25, 0.25, 1}
\definecolor{yellow}{rgb}{1, 1, 0.7}

\newcommand{\cmark}{\ding{51}}%
\newcommand{\xmark}{\ding{55}}%

\newcommand*{\x}{\mathsf{x}\mskip1mu}

\newcommand{\modelname}{$k$-planes}
\newcommand{\Modelname}{$K$-planes}

\crefname{section}{Sec.}{Secs.}
\Crefname{section}{Section}{Sections}
\Crefname{table}{Table}{Tables}
\crefname{table}{Tab.}{Tabs.}

\newcommand{\greencheck}{{\color{green}\cmark}}
\newcommand{\redcross}{{\color{red}\xmark}}

\makeatletter
\renewcommand{\paragraph}{%
	\@startsection{paragraph}{4}%
	{\z@}{0.65ex \@plus 1ex \@minus .2ex}{-1em}% used to be 3.25ex
	{\normalfont \normalsize \bfseries }%
}
\makeatother
\def\cvprPaperID{10239}
\def\confName{CVPR}
\def\confYear{2023}
\newcommand\blfootnote[1]{%
  \begingroup
  \renewcommand\thefootnote{}\footnote{#1}%
  \addtocounter{footnote}{-1}%
  \endgroup
}

\begin{document}

\title{$K$-Planes:  Explicit Radiance Fields in Space, Time, and Appearance}

\author{Sara Fridovich-Keil*\\
UC Berkeley\\
{\tt\small sfk@berkeley.edu}
\and
Giacomo Meanti*\\
Istituto Italiano di Tecnologia\\
{\tt\small giacomo.meanti@iit.it}
\and
Frederik Rahb\ae k Warburg\\
Technical University of Denmark\\
{\tt\small frwa@dtu.dk}
\and
Benjamin Recht\\
UC Berkeley\\
{\tt\small brecht@berkeley.edu}
\and
Angjoo Kanazawa\\
UC Berkeley\\
{\tt\small kanazawa@berkeley.edu}
}

\maketitle

\begin{strip}\centering
\vspace*{-4em}  % otherwise abstract overflows in col 2.
\captionsetup{type=figure}
\includegraphics[width=\textwidth]{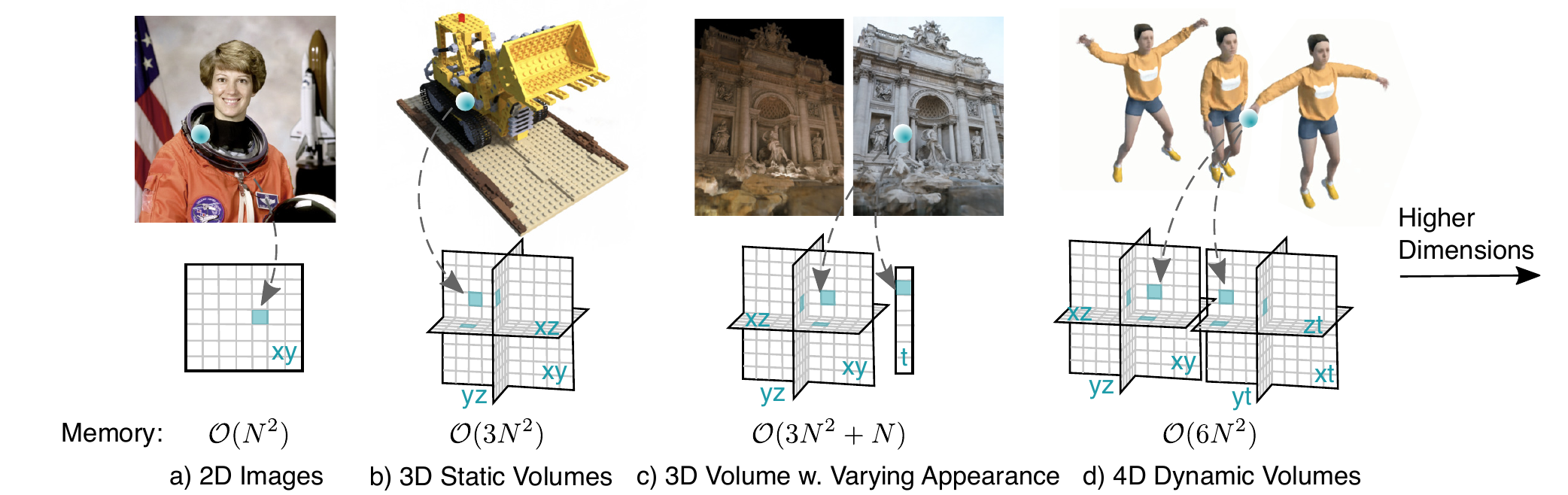}
\captionof{figure}{\textbf{Planar factorization of $d$-dimensional spaces.} We propose a simple planar factorization for volumetric rendering that naturally extends to arbitrary-dimensional spaces, and that scales gracefully with dimension in both optimization time and model size. We show the advantages of our approach on 3D static volumes, 3D photo collections with varying appearances, and 4D dynamic videos.
\label{fig:teaser}}
\end{strip}

\blfootnote{* equal contribution}

%%%%%%%%% ABSTRACT

\vspace{-10pt}

\begin{abstract}
We introduce $k$-planes, a white-box model for radiance fields in arbitrary dimensions. Our model uses $\binom{d}{2}$ (``$d$-\textit{choose}-$2$'') planes to represent a $d$-dimensional scene, providing a seamless way to go from static ($d=3$) to dynamic ($d=4$) scenes. This planar factorization makes adding dimension-specific priors easy, e.g. temporal smoothness and multi-resolution spatial structure, and induces a natural decomposition of static and dynamic components of a scene. We use a linear feature decoder with a learned color basis that yields similar performance as a nonlinear black-box MLP decoder. Across a range of synthetic and real, static and dynamic, fixed and varying appearance scenes, $k$-planes yields competitive and often state-of-the-art reconstruction fidelity with low memory usage, achieving $1000\x$ compression over a full 4D grid, and fast optimization with a pure PyTorch implementation. For video results and code, please see \url{https://sarafridov.github.io/K-Planes}.

\vspace{-1pt}

\end{abstract}

\vspace{-10pt}

\section{Introduction}\label{sec:intro}

Recent interest in dynamic radiance fields demands representations of 4D volumes. However, storing a 4D volume directly is prohibitively expensive due to the curse of dimensionality. Several approaches have been proposed to factorize 3D volumes for static radiance fields, but these do not easily extend to higher dimensional volumes.

We propose a factorization of 4D volumes that is simple, interpretable, compact, and yields fast training and rendering. Specifically, we use six planes to represent a 4D volume, where the first three represent space and the last three represent space-time changes, as illustrated in \cref{fig:teaser}(d). This decomposition of space and space-time makes our model interpretable, \ie dynamic objects are clearly visible in the space-time planes, whereas static objects only appear in the space planes. 
This interpretability enables dimension-specific priors in time and space.

More generally, our approach yields a straightforward, prescriptive way to select a factorization of any dimension with 2D planes.
For a $d$-dimensional space, we use $k = \binom{d}{2}$ (``$d$-\textit{choose}-$2$'') \emph{k-planes}, which represent every pair of dimensions --- for example, our model uses $\binom{4}{2}=6$ \emph{hex-planes} in 4D and reduces to $\binom{3}{2}=3$ \emph{tri-planes} in 3D. Choosing any other set of planes would entail either using more than $k$ planes and thus occupying unnecessary memory, or using fewer planes and thereby forfeiting the ability to represent some potential interaction between two of the $d$ dimensions. We call our model \modelname{}; \cref{fig:teaser} illustrates its natural application to both static and dynamic scenes.

Most radiance field models entail some black-box components with their use of MLPs. 
Instead, we seek a simple model whose functioning can be inspected and understood. 
We find two design choices to be fundamental in allowing \modelname{} to be a white-box model while maintaining reconstruction quality competitive with or better than previous black-box models~\cite{dynerf, dnerf}:
(1) Features from our \modelname{} are \emph{multiplied} together rather than added, as was done in prior work \cite{triplane, tensorf}, and (2) our linear feature decoder uses a learned basis for view-dependent color, enabling greater adaptivity including the ability to model scenes with variable appearance. We show that an MLP decoder can be replaced with this linear feature decoder only when the planes are multiplied, suggesting that the former is involved in both view-dependent color and determining spatial structure.

Our factorization of 4D volumes into 2D planes leads to a high compression level without relying on MLPs, using $200$ MB to represent a 4D volume whose direct representation at the same resolution would require more than $300$ GB, a compression rate of three orders of magnitude.
% a dynamic dataset weighing in at \SI{1.1}{\giga\byte}.
Furthermore, despite not using any custom CUDA kernels, \modelname{} trains orders of magnitude faster than prior implicit models and on par with concurrent hybrid models.

In summary, we present the first white-box, interpretable model capable of representing radiance fields in arbitrary dimensions, including static scenes, dynamic scenes, and scenes with variable appearance. Our \modelname{} model achieves competitive performance across reconstruction quality, model size, and optimization time across these varied tasks, without any custom CUDA kernels.

\section{Related Work}
\label{sec:related_work}

\Modelname{} is an interpretable, explicit model applicable to static scenes, scenes with varying appearances, and dynamic scenes, with compact model size and fast optimization time. Our model is the first to yield all of these attributes, as illustrated in \cref{tab:relatedwork}. 
We further highlight that \modelname{} satisfies this in a simple framework that naturally extends to arbitrary dimensions.  

\begin{table}[ht!]
    \centering
    \resizebox{1\linewidth}{!}{
    \begin{tabular}{l|cccccc}
        \toprule
         &  \rotatebox[origin=l]{90}{Static} & \rotatebox[origin=l]{90}{Appearance} & \rotatebox[origin=l]{90}{Dynamic} & \rotatebox[origin=l]{90}{Fast} & \rotatebox[origin=l]{90}{Compact} & \rotatebox[origin=l]{90}{Explicit} \\ \midrule
         NeRF & \greencheck         &    \redcross       &     \redcross      &   \redcross & \greencheck & \redcross      \\
         NeRF-W &     \greencheck     &       \greencheck    & \redcross          &    \redcross   & \greencheck & \redcross      \\
         DVGO &     \greencheck     &  \redcross         &    \redcross       & \greencheck      &    \redcross  & \redcross \\
         Plenoxels &     \greencheck     &  \redcross         &    \redcross       & \greencheck      &    \redcross  & \greencheck \\
         Instant-NGP, TensoRF  &         \greencheck     &  \redcross         &    \redcross       & \greencheck      &    \greencheck  & \redcross$^1$ \\
         DyNeRF, D-NeRF &       --     &  \redcross         &    \greencheck       & \redcross      &    \greencheck   & \redcross \\
         TiNeuVox, Tensor4D &      --     &  \redcross         &    \greencheck       & \greencheck      &    \greencheck & \redcross  \\
         MixVoxels, V4D &      --     &  \redcross         &    \greencheck       & \greencheck      &    \redcross & \redcross  \\
         NeRFPlayer &      --     &  \redcross         &    \greencheck       & \greencheck      &    \greencheck$^2$  & \redcross \\ \midrule
         \Modelname{} hybrid (Ours) & \greencheck  &  \greencheck &  \greencheck  &  \greencheck  & \greencheck & \redcross \\ 
         \Modelname{} explicit (Ours) & \greencheck  &  \greencheck &  \greencheck  &  \greencheck  & \greencheck & \greencheck \\ \bottomrule
    \end{tabular}}
    \vspace{-0.1cm}
    \begin{flushleft}
    {\footnotesize $^1$ TensoRF offers both hybrid and explicit versions, with a small quality gap}
    {\footnotesize $^2$ NerfPlayer offers models at different sizes, the smallest of which has $<100$ million parameters but the largest of which has $>300$ million parameters}
    \end{flushleft}
    \caption{\textbf{Related work overview.} The \modelname{} model works for a diverse set of scenes and tasks (static, varying appearance, and dynamic). It has a low memory usage (compact) and fast training and inference time (fast). Here ``fast'' includes any model that can optimize within a few ($<6$) hours on a single GPU, and ``compact'' denotes models that use less than roughly 100 million parameters. ``Explicit'' denotes white-box models that do not rely on MLPs.}
    \label{tab:relatedwork}
\end{table}

\begin{figure*}
    \centering
    \includegraphics[width=\textwidth]{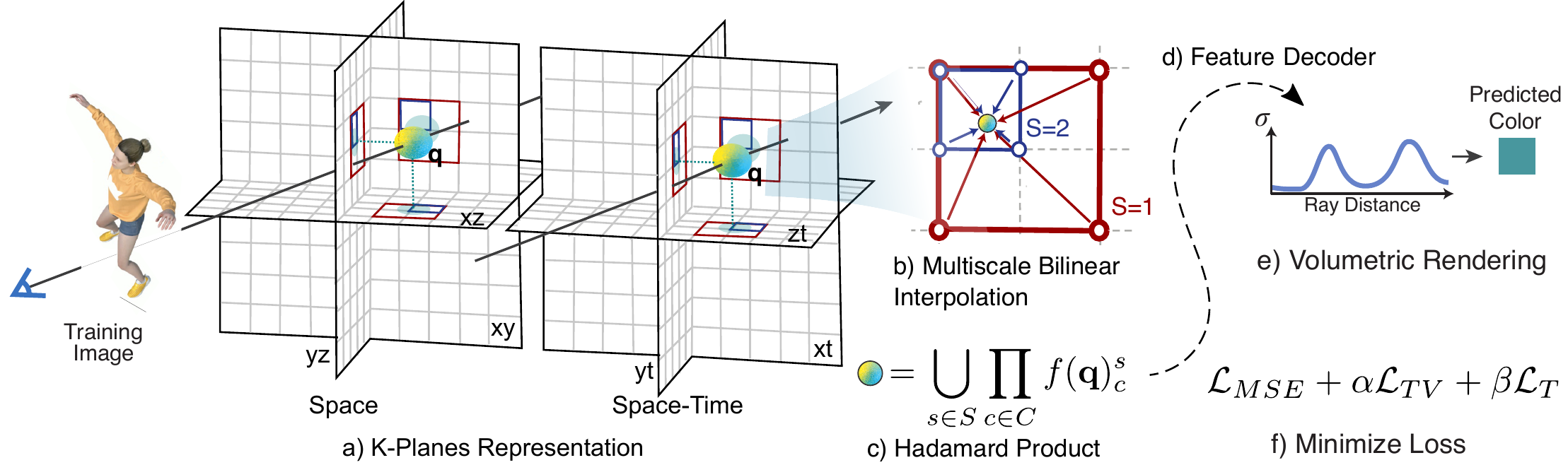}
    \caption{\textbf{Method overview.} (a) Our \modelname{} representation factorizes 4D dynamic volumes into six planes, three for space and three for spatiotemporal variations. To obtain the value of a 4D point $\textbf{q}=(x,y,z,t)$, we first project the point into each plane, in which we (b) do multiscale bilinear interpolation. (c) The interpolated values are multiplied and then concatenated over the $S$ scales. (d) These features are decoded either with a small MLP or our explicit linear decoder. (e) We follow the standard volumetric rendering formula to predict ray color and density. The model is optimized by (f) minimizing the reconstruction loss with simple regularization in space and time.}
    \label{fig:overview}
\end{figure*}

\paragraph{Spatial decomposition.}
NeRF\cite{nerf} proposed a fully implicit model with a large neural network queried many times during optimization, making it slow and essentially a black-box. 
Several works have used geometric representations to reduce the optimization time. Plenoxels~\cite{plenoxels} proposed a fully explicit model with trilinear interpolation in a 3D grid, which reduced the optimization time from hours to a few minutes. 
However, their explicit grid representation of 3D volumes, and that of DVGO~\cite{dvgo}, grows exponentially with dimension, making it challenging to scale to high resolution and completely intractable for 4D dynamic volumes.

Hybrid methods \cite{dvgo, ingp, tensorf} retain some explicit geometric structure, often compressed by a spatial decomposition, alongside a small MLP feature decoder. Instant-NGP \cite{ingp} proposed a multiresolution voxel grid encoded implicitly via a hash function, allowing fast optimization and rendering with a compact model. TensoRF \cite{tensorf} achieved similar model compression and speed by replacing the voxel grid with a tensor decomposition into planes and vectors. In a generative setting, EG3D \cite{triplane} proposed a similar spatial decomposition into three planes, whose values are added together to represent a 3D volume. 

Our work is inspired by the explicit modeling of Plenoxels as well as these spatial decompositions, particularly the triplane model of \cite{triplane}, the tensor decomposition of \cite{tensorf}, and the multiscale grid model of \cite{ingp}. We also draw inspiration from Extreme MRI \cite{extreme_mri}, which uses a multiscale low-rank decomposition to represent 4D dynamic volumes in magnetic resonance imaging. These spatial decomposition methods have been shown to offer a favorable balance of memory efficiency and optimization time for static scenes. However, it is not obvious how to extend these factorizations to 4D volumes in a memory-efficient way. \Modelname{} defines a unified framework that enables efficient and interpretable factorizations of 3D and 4D volumes and trivially extends to even higher dimensional volumes.

\paragraph{Dynamic volumes.}
Applications such as Virtual Reality (VR) and Computed Tomography (CT) often require the ability to reconstruct 4D volumes. Several works have proposed extensions of NeRF to dynamic scenes. The two most common schemes are (1) modeling a deformation field on top of a static \emph{canonical} field~\cite{dnerf, nrnerf, nerfies, Du4d2021, yuan2021star, tineuvox, nsff}, or (2) directly learning a radiance field conditioned on time\cite{xian2021space, nsff, gao2021, dynerf, hypernerf}. The former makes decomposing static and dynamic components easy \cite{yuan2021star, d2nerf}, but struggles with changes in scene topology (e.g. when a new object appears), while the latter makes disentangling static and dynamic objects hard.
A third strategy is to choose a representation of 3D space and repeat it at each timestep (e.g. NeRFPlayer \cite{nerfplayer}), resulting in a model that ignores space-time interactions and can become impractically large for long videos. 
% A third strategy is to choose a compact 3D spatial representation and essentially repeat it at each timestep (e.g. NeRFPlayer \cite{nerfplayer}), resulting in a model that ignores interactions between space and time and can become impractically large for long videos. 

Further, some of these models are fully implicit \cite{dnerf, dynerf} and thus suffer from extremely long training times (e.g. DyNeRF used 8 GPUs for 1 week to train a single scene), as well as being completely black-box. Others use partially explicit decompositions for video \cite{tineuvox, nvgd, mixvoxels, v4d, tensor4d, devrf, neuralvolumes, nerfplayer}, usually combining some voxel or spatially decomposed feature grid with one or more MLP components for feature decoding and/or representing scene dynamics. Most closely related to \modelname{} is Tensor4D~\cite{tensor4d}, which uses 9 planes to decompose 4D volumes. \Modelname{} is less redundant (\eg Tensor4D includes two $yt$ planes), does not rely on multiple MLPs, and offers a simpler factorization that naturally generalizes to static and dynamic scenes.
Our method combines a fully explicit representation with a built-in decomposition of static and dynamic components, the ability to handle arbitrary topology and lighting changes over time, fast optimization, and compactness.

\paragraph{Appearance embedding.}
Reconstructing large environments from photographs taken with varying illumination is another domain in which implicit methods have shown appealing results, but hybrid and explicit approaches have not yet gained a foothold. NeRF-W~\cite{martinbrualla2020nerfw} was the first to demonstrate photorealistic view synthesis in such environments. They augment a NeRF-based model with a learned global appearance code per frame, enabling it to explain away changes in appearance, such as time of day. With Generative Latent Optimization (GLO)~\cite{Piotr2017latentspaceoptim}, these appearance codes can further be used to manipulate the scene appearance by interpolation in the latent appearance space. Block-NeRF~\cite{blocknerf2022tancik} employs similar appearance codes.

We show that our \modelname{} representation can also effectively reconstruct these unbounded environments with varying appearance.
We similarly extend our model -- either the learned color basis in the fully explicit version, or the MLP decoder in the hybrid version -- with a global appearance code to disentangle global appearance from a scene without affecting geometry. To the best of our knowledge, ours is both the first fully explicit and the first hybrid method to successfully reconstruct these challenging scenes.

\section{K-planes model}
\label{sec:method}

We propose a simple and interpretable model for representing scenes in arbitrary dimensions. 
Our representation yields low memory usage and fast training and rendering. 
The \modelname{} factorization, illustrated in \cref{fig:overview}, models a $d$-dimensional scene using $k = \binom{d}{2}$ planes representing every combination of two dimensions. For example, for static 3D scenes, this results in \emph{tri-planes} with $\binom{3}{2} = 3$ planes representing $xy$, $xz$, and $yz$. For dynamic 4D scenes, this results in \emph{hex-planes}, with $\binom{4}{2} = 6$ planes including the three space-only planes and three space-time planes $xt$, $yt$, and $zt$. Should we wish to represent a 5D space, we could use $\binom{5}{2} = 10$ \emph{deca-planes}. In the following section, we describe the 4D instantiation of our \modelname{} factorization.

\subsection{Hex-planes}\label{sec:kplane}

The hex-planes factorization uses six planes. We refer to the space-only planes as $\textbf{P}_{xy}$, $\textbf{P}_{xz}$, and $\textbf{P}_{yz}$, and the space-time planes as $\textbf{P}_{xt}$, $\textbf{P}_{yt}$, and $\textbf{P}_{zt}$. Assuming symmetric spatial and temporal resolution $N$ for simplicity of illustration, each of these planes has shape $N \x N \x M$, where $M$ is the size of stored features that capture the density and view-dependent color of the scene. 

We obtain the features of a 4D coordinate $\bm{q} = (i, j, k, \tau)$ by normalizing its entries between $\left[0, N\right)$ and projecting it onto these six planes 
\begin{equation}\label{eq:projection}
    f(\bm{q})_c = \psi\big(\textbf{P}_c, \pi_c(\bm{q})\big),
\end{equation}
where $\pi_c$ projects $\bm{q}$ onto the $c$'th plane and $\psi$ denotes bilinear interpolation of a point into a regularly spaced 2D grid. We repeat \cref{eq:projection} for each plane $c\in C$ to obtain feature vectors $f(\bm{q})_c$. We combine these features over the six planes using the Hadamard product (elementwise multiplication) to produce a final feature vector of length $M$
\begin{equation}\label{eq:contraction}
    f(\bm{q}) = \prod_{c\in C} f(\bm{q})_c.
\end{equation}
These features will be decoded into color and density using either a linear decoder or an MLP, described in \cref{sec:decoder}.

\begin{figure}[ht!]
\begin{minipage}[]{1\linewidth}
  \centering
  \includegraphics[width=0.45\linewidth]{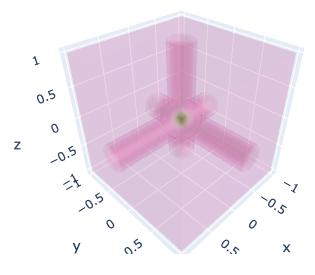}
  \includegraphics[width=0.45
  \linewidth]{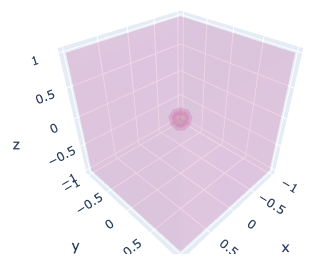}
\end{minipage}
\caption{\textbf{Addition versus Hadamard product.} Elementwise addition of plane features (left) compared to multiplication (right), in a triplane example. A single entry in each plane is positive and the rest are zero, selecting a single 3D point by multiplication but producing intersecting lines by addition. This selection ability of multiplication improves the expressivity of our explicit model.}
\label{fig:multiplication}
\end{figure}

\paragraph{Why Hadamard product?}
\label{sec:hadamard} \looseness=-1
In 3D, \modelname{} reduces to the tri-plane factorization, which is similar to ~\cite{triplane} except that the elements are multiplied. A natural question is why we multiply rather than add, as has been used in prior work with tri-plane models \cite{triplane, convtriplane}.
\cref{fig:multiplication} illustrates that combining the planes by multiplication allows \modelname{} to produce spatially localized signals, which is not possible with addition.

This selection ability of the Hadamard product produces substantial rendering improvements for linear decoders and modest improvement for MLP decoders, as shown in \cref{tab:multiplication}. This suggests that the MLP decoder is involved in both view-dependent color and determining spatial structure. The Hadamard product relieves the feature decoder of this extra task and makes it possible to reach similar performance using a linear decoder solely responsible for view-dependent color.

\renewcommand{\tabcolsep}{6pt}
\begin{table}[ht]
  \centering
  \resizebox{\linewidth}{!}{%
  \begin{tabular}{rlccc}
    \multicolumn{5}{c}{} \\
    \toprule 
    Plane Combination & & Explicit  & Hybrid & \# params $\downarrow$ \\
    \cmidrule(){1-1} \cmidrule(){3-5} 
    % Multiplication &&	36.36 &	36.97 &	34M \\
    % Addition	&& 30.59 &	36.73 &	34M \\
    Multiplication && 35.29 & 35.75 & 33M \\
    Addition && 28.78 & 34.80 & 33M \\
    \bottomrule
  \end{tabular}}
  \caption{\textbf{Ablation study over Hadamard product.} Multiplication of plane features yields a large improvement in PSNR $\uparrow$ for our explicit model, whereas our hybrid model can use its MLP decoder to partially compensate for the less expressive addition of planes. This experiment uses the static \emph{Lego} scene \cite{nerf} with $3$ scales: $128$, $256$, and $512$, and $32$ features per scale.}
  \label{tab:multiplication}
\end{table}

\subsection{Interpretability} \label{sec:regularization}
The separation of space-only and space-time planes makes the model interpretable and enables us to incorporate dimension-specific priors. For example, if a region of the scene never moves, its temporal component will always be $1$ (the multiplicative identity), thereby just using the features from the space planes. This offers compression benefits since a static region can easily be identified and compactly represented. Furthermore, the space-time separation improves interpretability, \ie we can track the changes in time by visualizing the elements in the time-space planes that are not $1$. This simplicity, separation, and interpretability make adding priors straightforward.

\paragraph{Multiscale planes.}
To encourage spatial smoothness and coherence, our model contains multiple copies at different spatial resolutions, for example $64$, $128$, $256$, and $512$.
% , to encode a prior of spatial smoothness and coherence.
Models at each scale are treated separately, and the $M$-dimensional feature vectors from different scales are concatenated together before being passed to the decoder. The red and blue squares in \cref{fig:overview}a-b illustrate bilinear interpolation with multiscale planes. Inspired by the multiscale hash mapping of Instant-NGP\cite{ingp}, this representation efficiently encodes spatial features at different scales, allowing us to reduce the number of features stored at the highest resolution and thereby further compressing our model.  Empirically, we do not find it necessary to represent our time dimension at multiple scales. 

\paragraph{Total variation in space.}
Spatial total variation regularization encourages sparse gradients (with L1 norm) or smooth gradients (with L2 norm), encoding priors over edges being either sparse or smooth in space. We encourage this in 1D over the spatial dimensions of each of our space-time planes and in 2D over our space-only planes:
\begin{equation}
    \mathcal{L}_{TV}(\textbf{P}) = \frac{1}{\lvert C\rvert n^2} \sum_{c,i,j} 
        \big(\lVert \textbf{P}_c^{i,j} - \textbf{P}_c^{i-1,j}\rVert_2^2 +
         \lVert \textbf{P}_c^{i,j} - \textbf{P}_c^{i,j-1}\rVert_2^2 \big),
\end{equation}
where $i,j$ are indices on the plane's resolution. Total variation is a common regularizer in inverse problems and was used in Plenoxels \cite{plenoxels} and TensoRF \cite{tensorf}. We use the L2 version in our results, though we find that either L2 or L1 produces similar quality.

\paragraph{Smoothness in time.}
We encourage smooth motion with a 1D Laplacian (second derivative) filter
\begin{equation}
    \mathcal{L}_{smooth}(\textbf{P}) = \frac{1}{\lvert C\rvert n^2} \sum_{c,i,t} 
        \lVert \textbf{P}_c^{i,t-1} -2\textbf{P}_c^{i,t} + \textbf{P}_c^{i,t+1}\rVert_2^2,
\end{equation}
to penalize  sharp ``acceleration'' over time. We only apply this regularizer on the time dimension of our space-time planes. Please see the appendix for an ablation study.

\paragraph{Sparse transients.}
We want the static part of the scene to be modeled by the space-only planes. We encourage this separation of space and time by initializing the features in the space-time planes as $1$ (the multiplicative identity) and using an $\ell_1$ regularizer on these planes during training:
\begin{equation}
\mathcal{L}_{sep}(\textbf{P}) = \sum_c \lVert \bm{1} - \textbf{P}_{c} \rVert_1, \qquad c\in\{xt, yt, zt\}.
\end{equation}
In this way, the space-time plane features of the \modelname{} decomposition will remain fixed at $1$ if the corresponding spatial content does not change over time.

\subsection{Feature decoders}\label{sec:decoder}

We offer two methods to decode the $M$-dimensional temporally- and spatially-localized feature vector $f(\bm{q})$ from \cref{eq:contraction} into density, $\sigma$, and view-dependent color, $\bm{c}$. 

\paragraph{Learned color basis: a linear decoder and explicit model.}
Plenoxels \cite{plenoxels}, Plenoctrees \cite{plenoctrees}, and TensoRF \cite{tensorf} proposed models where spatially-localized features are used as coefficients of the spherical harmonic (SH) basis, to describe view-dependent color. Such SH decoders can give both high-fidelity reconstructions and enhanced interpretability compared to MLP decoders. However, SH coefficients are difficult to optimize, and their expressivity is limited by the number of SH basis functions used (often limited 2nd degree harmonics, which produce blurry specular reflections).%. Often only harmonics up to degree two are used, thereby producing low-frequency (blurry) specular reflections.  

Instead, we replace the SH functions with a learned basis, retaining the interpretability of treating features as coefficients for a linear decoder yet increasing the expressivity of the basis and allowing it to adapt to each scene, as was proposed in NeX \cite{wizadwongsa2021nex}. We represent the basis using a small MLP that maps each view direction $\bm{d}$ to red $b_R(\bm{d}) \in \mathbb{R}^M$, green $b_G(\bm{d}) \in \mathbb{R}^M$, and blue $b_B(\bm{d}) \in \mathbb{R}^M$ \emph{basis vectors}. The MLP serves as an adaptive drop-in replacement for the spherical harmonic basis functions repeated over the three color channels. We obtain the color values

\begin{equation}\label{eq:learnedbasisrgbdecoder}
    \bm{c}(\bm{q}, \bm{d}) = \bigcup_{i \in \{R,G,B\}} f(\bm{q}) \cdot b_i(\bm{d}),
\end{equation}
where $\cdot$ denotes the dot product and $\cup$ denotes concatenation. Similarly, we use a learned basis $b_{\sigma} \in \mathbb{R}^M$, independent of the view direction, as a linear decoder for density:

\begin{equation}\label{eq:learnedbasissigmadecoder}
    \sigma(\bm{q}) = f(\bm{q}) \cdot b_{\sigma}.
\end{equation}
Predicted color and density values are finally forced to be in their valid range by applying the sigmoid to $\bm{c}(\bm{q}, \bm{d})$, and the exponential (with truncated gradient) to $\sigma(\bm{q})$.% through a nonlinear function to ensure they are in the valid ranges: we apply sigmoid to the color values to ensure they lie in $[0, 1]$, and exponential (with truncated gradient) to the density values to ensure they are nonnegative. 

\paragraph{MLP decoder: a hybrid model.}
Our model can also be used with an MLP decoder like that of Instant-NGP~\cite{ingp} and DVGO~\cite{dvgo}, turning it into a hybrid model. In this version, features are decoded by two small MLPs, one $g_{\sigma}$ that maps the spatially-localized features into density $\sigma$ and additional features $\hat f$, and another $g_{RGB}$ that maps $\hat{f}$ and %these additional features as well as  
the embedded view direction $\gamma(\bm{d})$ into RGB color

\begin{equation}\label{eq:mlpdecoder}
\begin{aligned}
    \sigma(\bm{q}), \hat f (\bm{q}) &= g_{\sigma}(f(\bm{q}))\\
    \bm{c}(\bm{q}, \bm{d}) &= g_{RGB}(\hat f (\bm{q}), \gamma(\bm{d})).
\end{aligned}
\end{equation}

As in the linear decoder case, the predicted density and color values are finally normalized via exponential and sigmoid, respectively.

\paragraph{Global appearance.}

We also show a simple extension of our \modelname{} model that enables it to represent scenes with consistent, static geometry viewed under varying lighting or appearance conditions. Such scenes appear in the Phototourism \cite{phototourism} dataset of famous landmarks photographed at different times of day and in different weather. To model this variable appearance, we augment \modelname{} with an $M$-dimensional vector for each training image $1, \dots, T$. % a matrix of shape $M \times T$, an $M$-vector of features for each training image index $1 \dots T$. 
Similar to NeRF-W \cite{martinbrualla2020nerfw}, we optimize this per-image feature vector and pass it as an additional input to either the MLP learned color basis $b_{R}, b_{G}, b_{B}$, in our explicit version, or to the MLP color decoder $g_{RGB}$, in our hybrid version, so that it can affect color but not geometry.

\newcommand{\plotcropship}[1]{%
\adjincludegraphics[trim={{0.6\width} {0.5\height} {0.3\width} {0.416\height}}, clip, width=\linewidth]{#1}%
}
\newcommand{\plotcropdog}[1]{%
\adjincludegraphics[trim={{0.45\width} {0.45\height} {0.45\width} {0.4655\height}}, clip, width=\linewidth]{#1}%
}
\begin{figure}[t]
  \centering
  \begin{subfigure}[b]{0.24\linewidth}
    \plotcropship{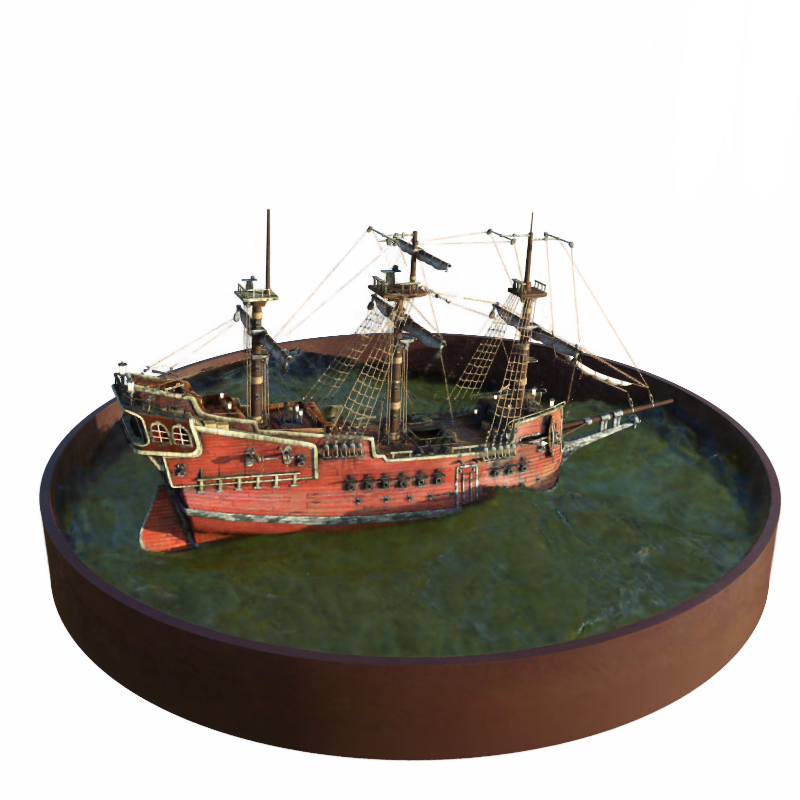}
    \plotcropdog{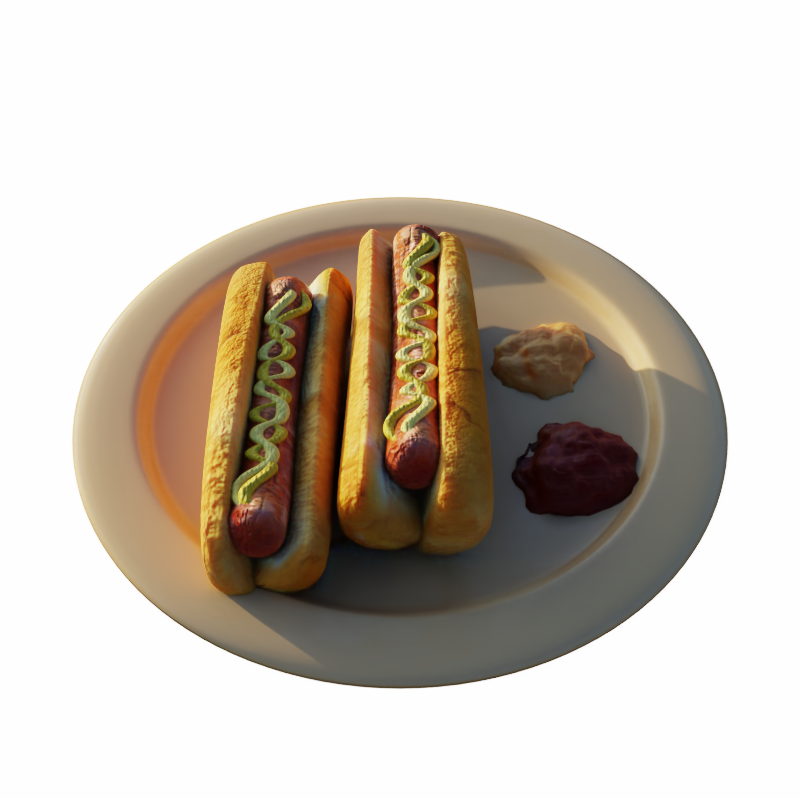}
    \caption{Ours-explicit}
  \end{subfigure}
  \begin{subfigure}[b]{0.24\linewidth}
    \plotcropship{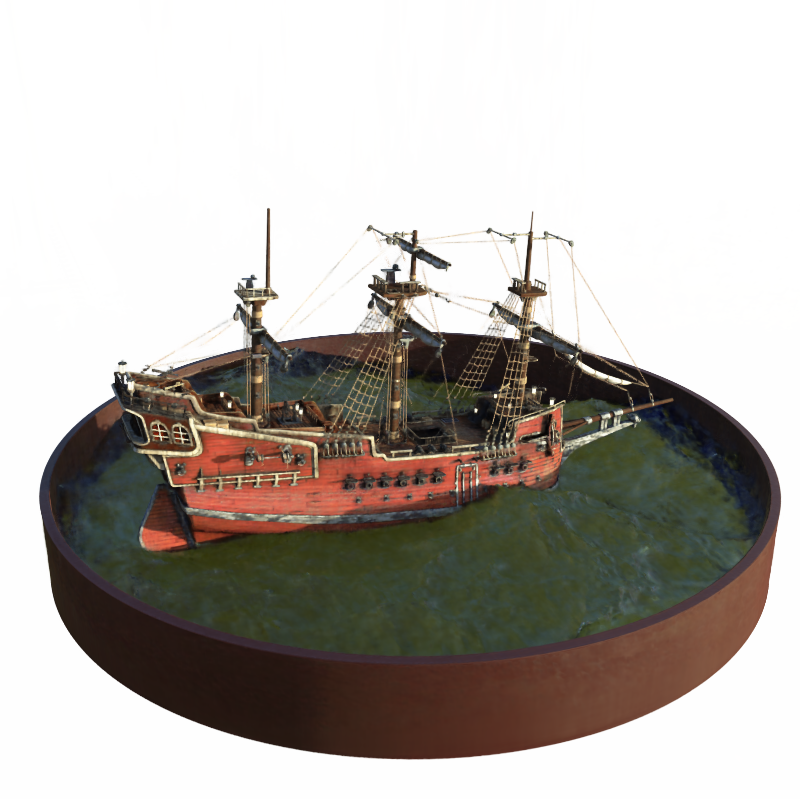}
    \plotcropdog{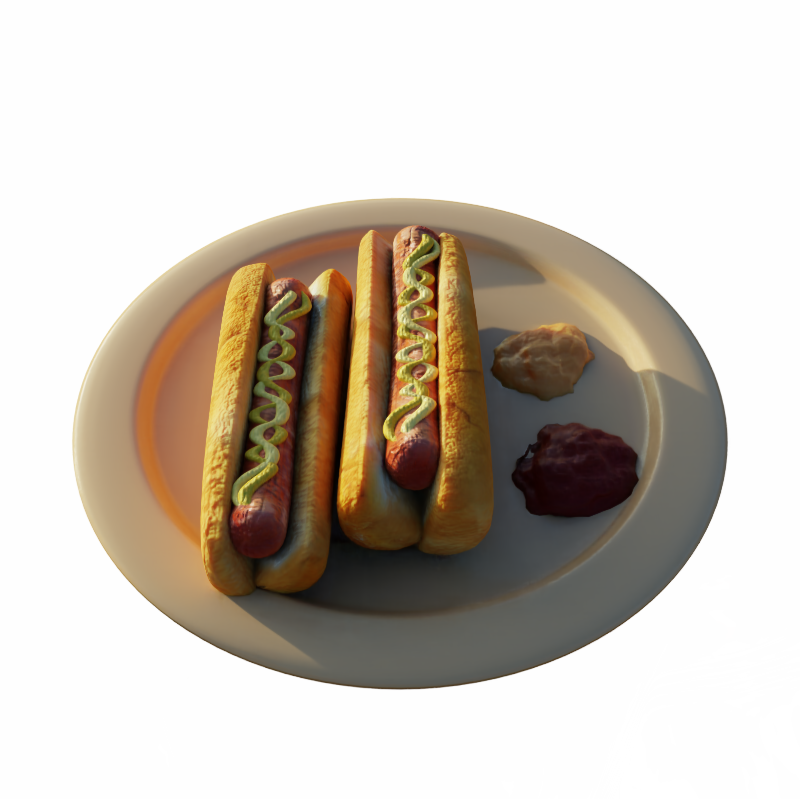}
    \caption{Ours-hybrid}
  \end{subfigure}
  \begin{subfigure}[b]{0.24\linewidth}
    \plotcropship{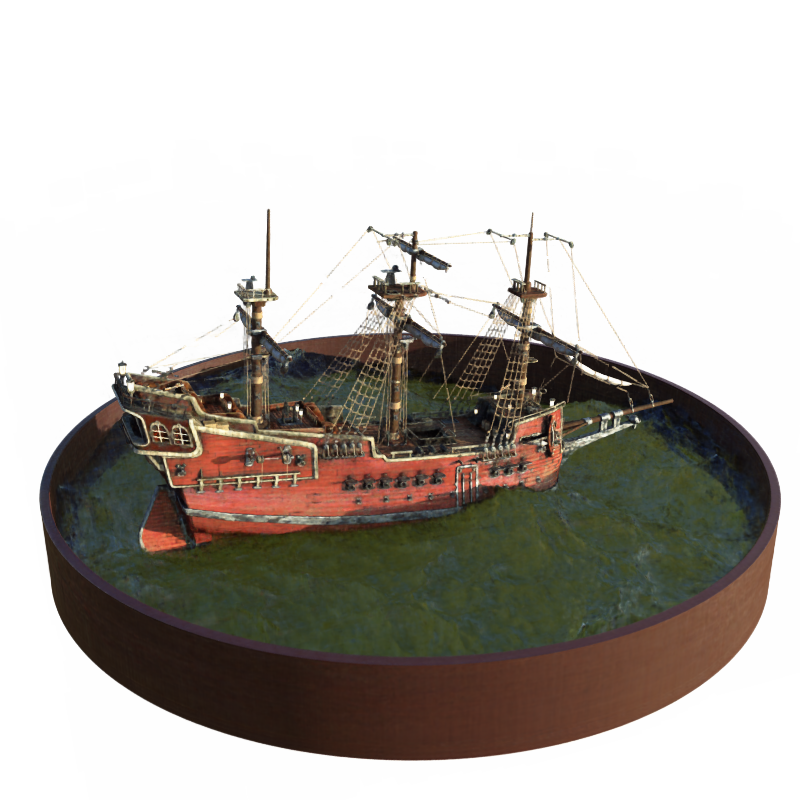}
    \plotcropdog{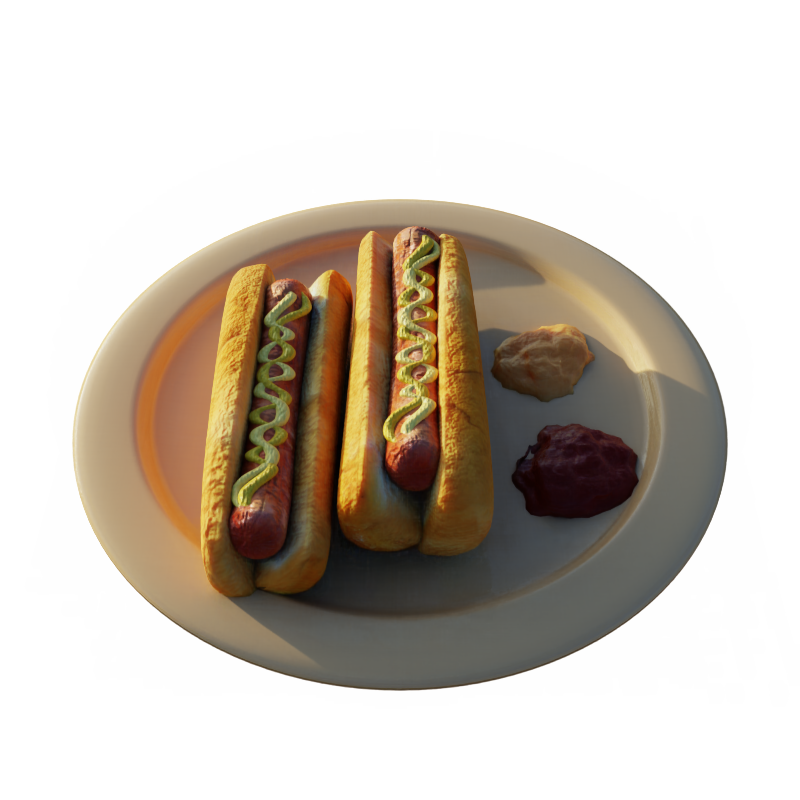}
    \caption{TensoRF}
  \end{subfigure}
  \begin{subfigure}[b]{0.24\linewidth}
    \plotcropship{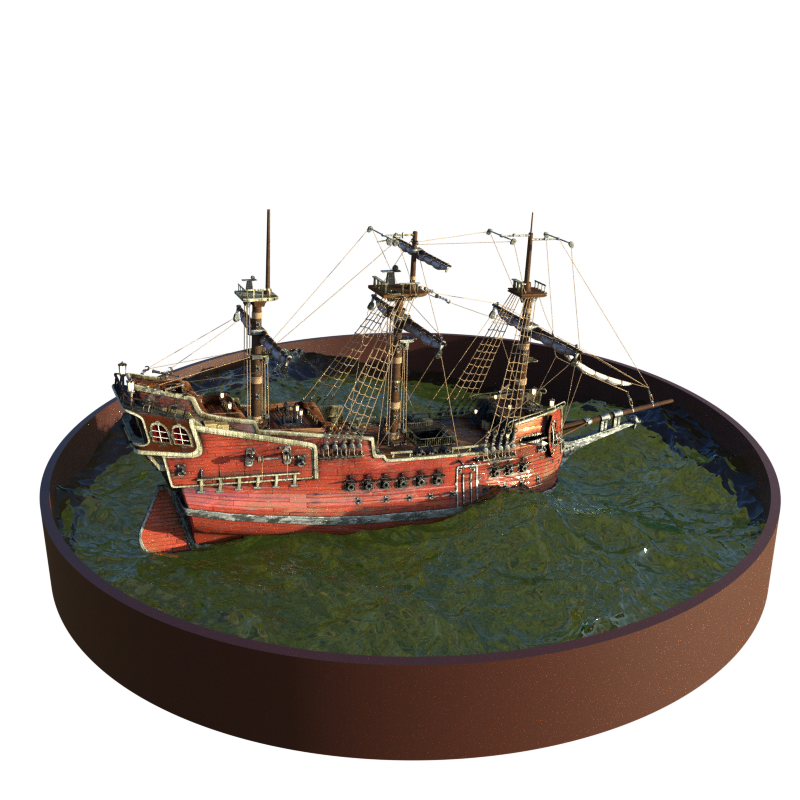}
    \plotcropdog{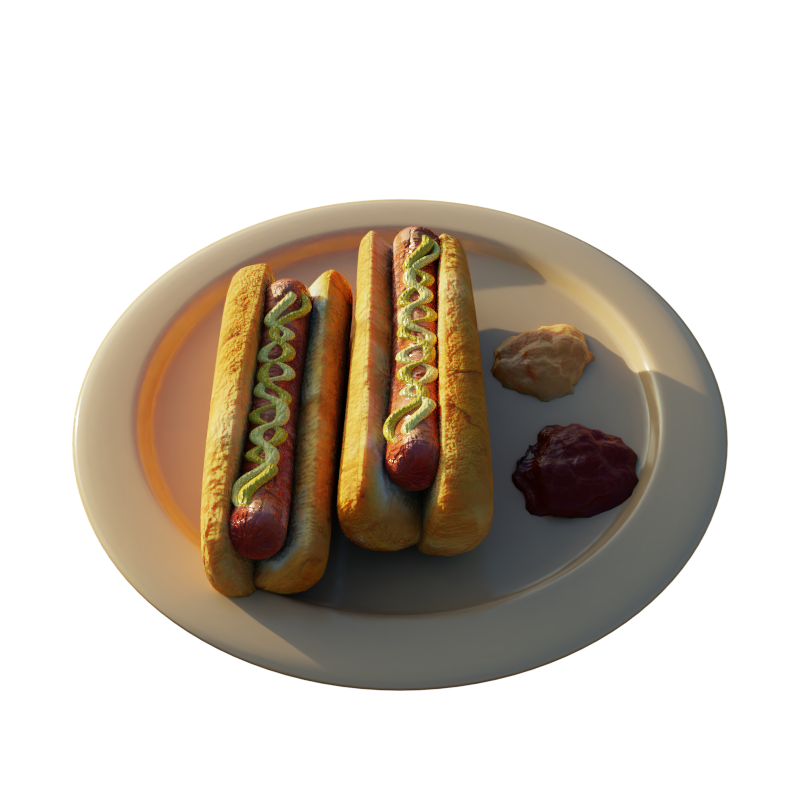}
    \caption{Ground truth}
  \end{subfigure}
  \caption{\textbf{Zoomed qualitative results on static NeRF scenes.} Visual comparison of \modelname{}, TensoRF \cite{tensorf}, and the ground truth, on \emph{ship} (top) and \emph{hotdog} (bottom).}
  \label{fig:synthetic-zoom}
\end{figure}

\subsection{Optimization details}
% Full details of our optimization procedure may be found in the released code. Here we specify three components we modify from prior work to improve flexibility and speed. \setlength{\parskip}{-7pt plus 2pt minus 2pt}

\paragraph{Contraction and normalized device coordinates.}%\looseness=-1
For forward-facing scenes, we apply normalized device coordinates (NDC) \cite{nerf} to better allocate our resolution while enabling unbounded depth. We also implement an $\ell_\infty$ version (rather than $\ell_2$) of the scene contraction proposed in Mip-NeRF 360 \cite{mipnerf360}, which we use on the unbounded Phototourism scenes.

\paragraph{Proposal sampling.}\looseness=-1
We use a variant of the proposal sampling strategy from Mip-NeRF 360 \cite{mipnerf360}, with a small instance of \modelname{} as density model. 
Proposal sampling works by iteratively refining density estimates along a ray, to allocate more points in the regions of higher density. We use a two-stage sampler, resulting in fewer samples that must be evaluated in the full model and in sharper details by placing those samples closer to object surfaces.
%Proposal sampling works by first using a fixed sampling strategy along each ray into a density model, and then adaptively choosing samples based on the densities of these initial samples. We use a two-stage proposal sampler, resulting in both fewer samples that must be evaluated in our full model and in sharper details by placing those samples closer to object surfaces. 
The density models used for proposal sampling are trained with the histogram loss~\cite{mipnerf360}.

\paragraph{Importance sampling.}\looseness=-1
For multiview dynamic scenes, we implement a version of the importance sampling based on temporal difference (IST) strategy from DyNeRF \cite{dynerf}. During the last portion of optimization, we sample training rays proportionally to the maximum variation in their color within 25 frames %, slightly less than one second 
before or after. This results in higher sampling probabilities in the dynamic region. We apply this strategy after the static scene has converged %through standard optimization 
with uniformly sampled rays. In our experiments, IST has only a modest impact on full-frame metrics but improves visual quality in the small dynamic region. Note that importance sampling cannot be used for monocular videos or datasets with moving cameras.

\section{Results}\label{sec:results}

We demonstrate the broad applicability of our planar decomposition via experiments in three domains: static scenes (both bounded $360^{\circ}$ and unbounded forward-facing), dynamic scenes (forward-facing multi-view and bounded $360^{\circ}$ monocular), and Phototourism scenes with variable appearance. For all experiments, we report the metrics PSNR (pixel-level similarity) and SSIM\footnote{Note that among prior work, some evaluate using an implementation of SSIM from MipNeRF \cite{mipnerf} whereas others use the scikit-image implementation, which tends to produce higher values. For fair comparison on each dataset we make a best effort to use the same SSIM implementation as the relevant prior work.}~\cite{ssim} (structural similarity), as well as approximate training time and number of parameters (in millions), in \cref{tab:results}. Blank entries in \cref{tab:results} denote baseline methods for which the corresponding information is not readily available. Full per-scene results may be found in the appendix.

\subsection{Static scenes}\label{sec:static}

\begin{figure}[t]
  \centering
  \begin{subfigure}[b]{0.24\linewidth}
    \includegraphics[width=\linewidth]{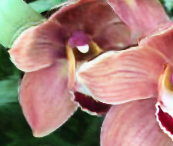}
    \includegraphics[width=\linewidth]{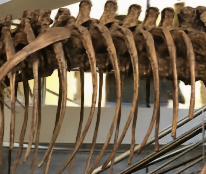}
    \caption{Ours-explicit}
  \end{subfigure}
  \begin{subfigure}[b]{0.24\linewidth}
    \includegraphics[width=\linewidth]{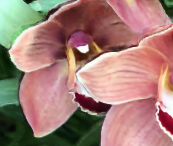}
    \includegraphics[width=\linewidth]{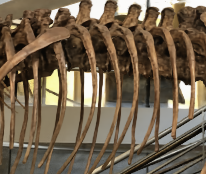}
    \caption{Ours-hybrid}
  \end{subfigure}
  \begin{subfigure}[b]{0.24\linewidth}
    \includegraphics[width=\linewidth]{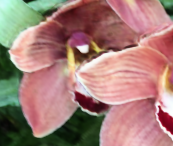}
    \includegraphics[width=\linewidth]{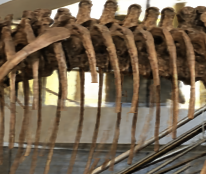}
    \caption{TensoRF}
  \end{subfigure}
  \begin{subfigure}[b]{0.24\linewidth}
    \includegraphics[width=\linewidth]{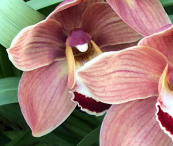}
    \includegraphics[width=\linewidth]{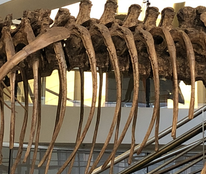}
    \caption{Ground truth}
  \end{subfigure}
  \caption{\textbf{Zoomed qualitative results on static LLFF scenes.} Visual comparison of \modelname{}, TensoRF~\cite{tensorf}, and the ground truth, on \emph{orchids} (top) and \emph{T-rex} (bottom).}
  \label{fig:llff-zoom}
\end{figure}

\begin{table}
 \renewcommand{\tabcolsep}{2pt}
 \resizebox{1\linewidth}{!}{
  \begin{threeparttable}
   \begin{tabular}{@{}llccccc@{}}
    \toprule
    && PSNR $\uparrow$ & SSIM $\uparrow$ & Train Time $\downarrow$ & \# Params $\downarrow$  \\ 
    \cmidrule{1-7}
    
    \multicolumn{7}{c}{NeRF~\cite{nerf} (static, synthetic)} \\
    \cmidrule(r){1-1} \cmidrule(l){3-7} 
    Ours-explicit              && 32.21 & 0.960 & 38 min & 33M \\
    Ours-hybrid                && 32.36 & 0.962 & 38 min & 33M \\
    % Ours-explicit              && 33.13 & 0.964 & 38 min & 34M             \\
    % Ours-hybrid                && 33.62 & 0.967 & 38 min & 34M             \\
    Plenoxels~\cite{plenoxels} && 31.71 & 0.958 & 11 min & $\sim$500M      \\
    TensoRF~\cite{tensorf}     && 33.14 & 0.963 & 17 min & 18M             \\
    I-NGP~\cite{ingp}          && 33.18 & -     & 5 min  & $\sim$ 16M      \\
    \cmidrule{1-7}
    
    \multicolumn{7}{c}{LLFF~\cite{llff} (static, real)}                    \\
    \cmidrule(r){1-1} \cmidrule(l){3-7}
    Ours-explicit              && 26.78 & 0.841 & 33 min & 19M             \\
    Ours-hybrid                && 26.92 & 0.847 & 33 min & 19M             \\
    Plenoxels                  && 26.29 & 0.839 & 24 min & $\sim$500M      \\
    TensoRF                    && 26.73 & 0.839 & 25 min & 45M             \\
    \cmidrule{1-7}

    \multicolumn{7}{c}{D-NeRF~\cite{dnerf} (dynamic, synthetic)}           \\
    \cmidrule(r){1-1} \cmidrule(l){3-7} 
    % Using half resolution and time smoothness 0.01
    Ours-explicit && 31.05 & 0.97 & 52 min & 37M & \\
    Ours-hybrid && 31.61 & 0.97 & 52 min & 37M & \\
    % Using full resolution and time smoothness 0.1
    % Ours-explicit              && 30.39 & 0.96 & 52 min & 37M &            \\
    % Ours-hybrid                && 30.84 & 0.96 & 52 min & 37M &            \\
    D-NeRF                     && 29.67 & 0.95 & 48 hrs & 1-3M &           \\
    TiNeuVox\cite{tineuvox} && 32.67 & 0.97 & 30 min & $\sim$12M &      \\
    V4D\cite{v4d}     && 33.72 & 0.98 & 4.9 hrs & 275M &        \\
    \cmidrule{1-7}

    \multicolumn{7}{c}{DyNeRF~\cite{dynerf} (dynamic, real)}               \\
    \cmidrule(r){1-1} \cmidrule(l){3-7}
    Ours-explicit              && 30.88 & 0.960 & 3.7 hrs & 51M            \\
    Ours-hybrid                && 31.63 & 0.964 & 1.8 hrs & 27M            \\
    DyNeRF~\cite{dynerf}       && $^1$29.58 & - & 1344 hrs  & 7M      \\
    LLFF~\cite{llff}           && $^1$23.24 & - & - & -               \\
    MixVoxels-L\cite{mixvoxels}&& 30.80 & 0.960 & 1.3 hrs & 125M\\
    \cmidrule{1-7}

    \multicolumn{7}{c}{Phototourism~\cite{phototourism} (variable appearance)} \\
    \cmidrule(r){1-1} \cmidrule(l){3-7}  
    Ours-explicit  && 22.25 & 0.859 & 35 min & 36M                         \\
    Ours-hybrid  && 22.92 & 0.877 & 35 min & 36M                           \\
    NeRF-W~\cite{martinbrualla2020nerfw} && 27.00 & 0.962 & 384 hrs & $\sim$2M           \\
    NeRF-W (public)\tnote{2}       && 19.70 & 0.764 & 164 hrs & $\sim$2M       \\
    LearnIt \cite{learnit}     && 19.26 & - & - & -                        \\
    \bottomrule
   \end{tabular}
  \end{threeparttable}
 }
 \\
 {
  \footnotesize %TiNeuVox uses half-resolution images for training and evaluation (V4D code is not yet public, so their resolution is unknown).
  $^1$\,DyNeRF and LLFF only report metrics on the \emph{flame salmon} video (the first 10 seconds); average performance may be higher as this is one of the more challenging videos.
  $^2$\,Open-source version \url{https://github.com/kwea123/nerf_pl/tree/nerfw} where we re-implemented test-time optimization as for \modelname{}.
  \par
 }
 \caption{\textbf{Results.} 
    Averaged metrics over all scenes in the respective datasets. Note that Phototourism scenes use MS-SSIM (multiscale structural similarity) instead of SSIM. 
   %\textit{First section:} average over the 8 synthetic static scenes from NeRF \cite{nerf}; 
   %\textit{Second section:} average over the 8 real forward-facing static scenes from LLFF \cite{llff}; 
   %\textit{Third section:} average over the 8 synthetic monocular ``teleporting camera'' dynamic scenes from D-NeRF \cite{dnerf}; 
   % \textit{Fourth section:} average over the 6 real multiview forward-facing dynamic scenes from DyNeRF \cite{dynerf}; 
   %\textit{Fifth section:} average over three scenes from the Phototourism dataset \cite{phototourism}; note these scenes use MS-SSIM (multiscale structural similarity) instead of SSIM. 
   \Modelname{} timings are based on a single NVIDIA A30 GPU. 
   Please see the appendix for per-scene results and the website for video reconstructions. 
 }\label{tab:results}
 \vspace{-2em}
\end{table}

\definecolor{col1}{HTML}{e0d291}
\definecolor{col2}{HTML}{d66079}
\newcommand\videozoomed[1]{
    % \raisebox{-0.5\height}{
    \begin{tikzpicture}[
    zoomboxarray,
    zoomboxes below,
    connect zoomboxes,
    zoombox paths/.append style={thick}]
        \node[image node]{\includegraphics[width=0.16\textwidth]{figures/video/#1.jpg}};
        \zoombox[magnification=3.5,color code=col1]{0.415,0.300}  
        \zoombox[magnification=6,color code=col2]{0.693,0.113} %
    \end{tikzpicture}
    % }
}
\newcommand\videozoomedmixvoxels[1]{
    % \raisebox{-0.5\height}{
    \begin{tikzpicture}[
    zoomboxarray,
    zoomboxes below,
    connect zoomboxes,
    zoombox paths/.append style={thick}]
        \node[image node]{\includegraphics[width=0.16\textwidth]{figures/video/#1.jpg}};
        \zoombox[magnification=3.5,color code=col1]{0.415,0.350}  
        \zoombox[magnification=6,color code=col2]{0.688,0.123} %
    \end{tikzpicture}
    % }
}
\begin{figure*}[t]\centering
    \def\arraystretch{1}
    \begin{tabular}{@{}c@{}c@{}c@{}c@{}c@{}c@{}}
    \videozoomed{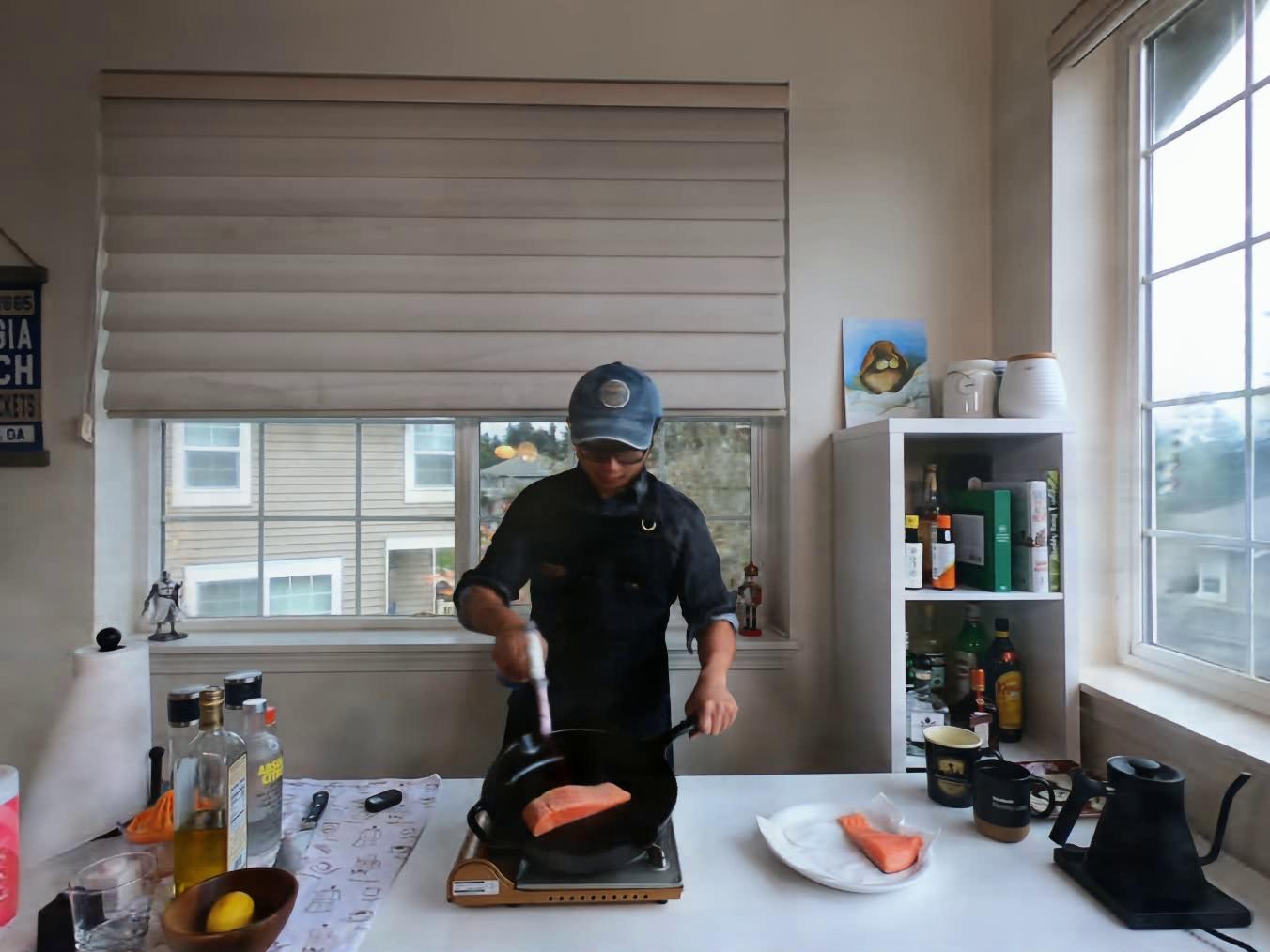} &
    \videozoomed{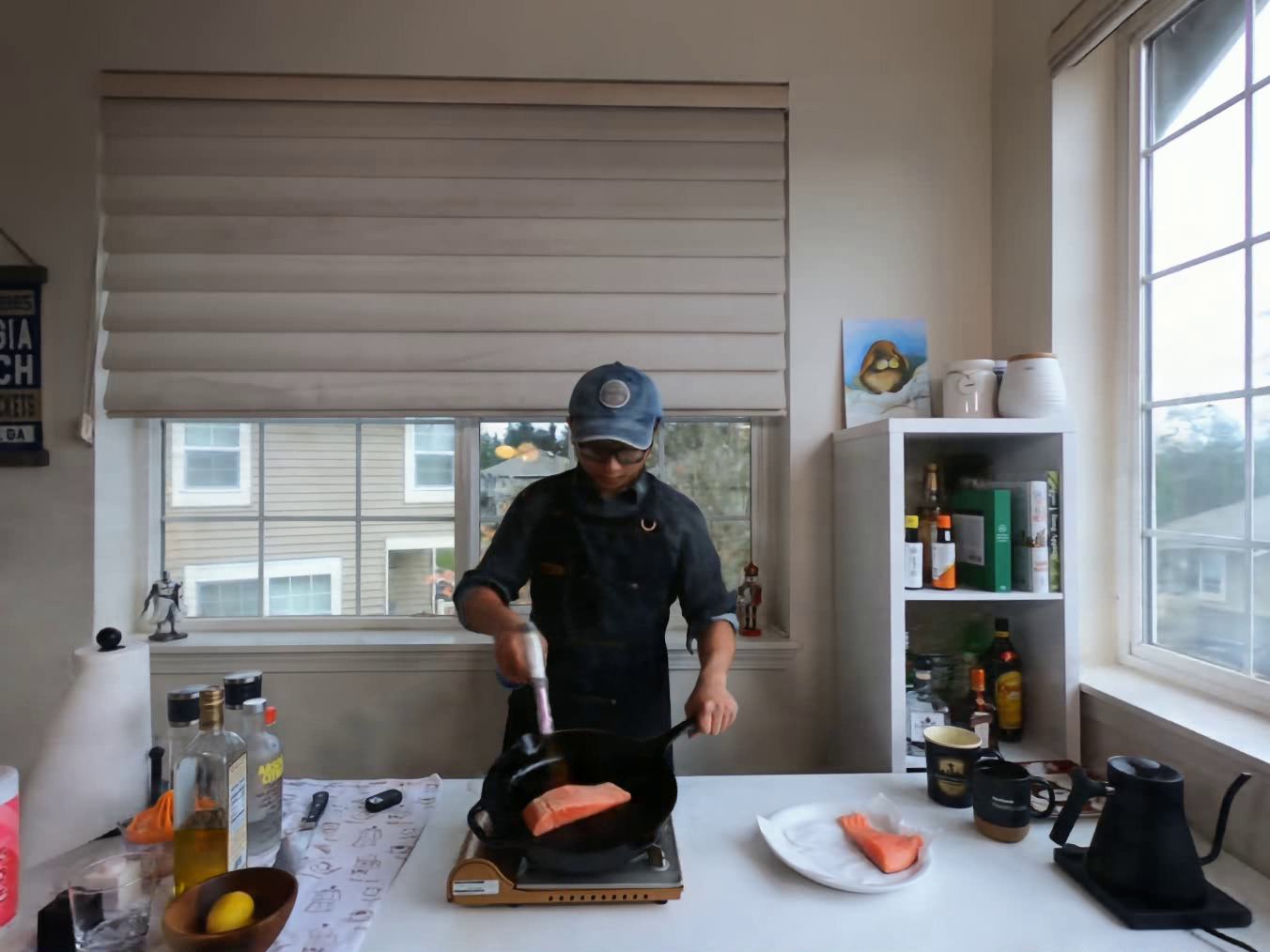} & 
    \videozoomed{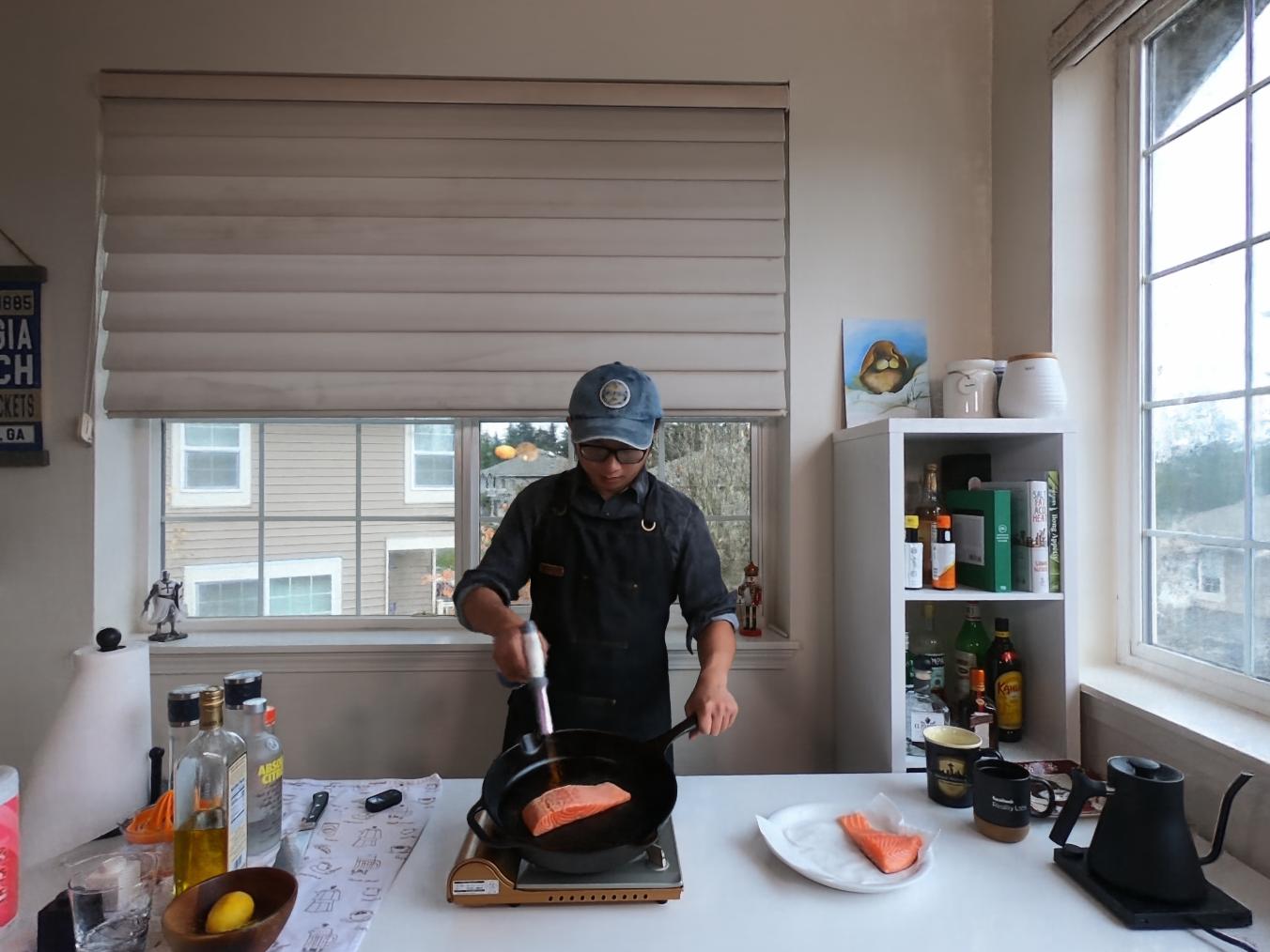} &
    \videozoomedmixvoxels{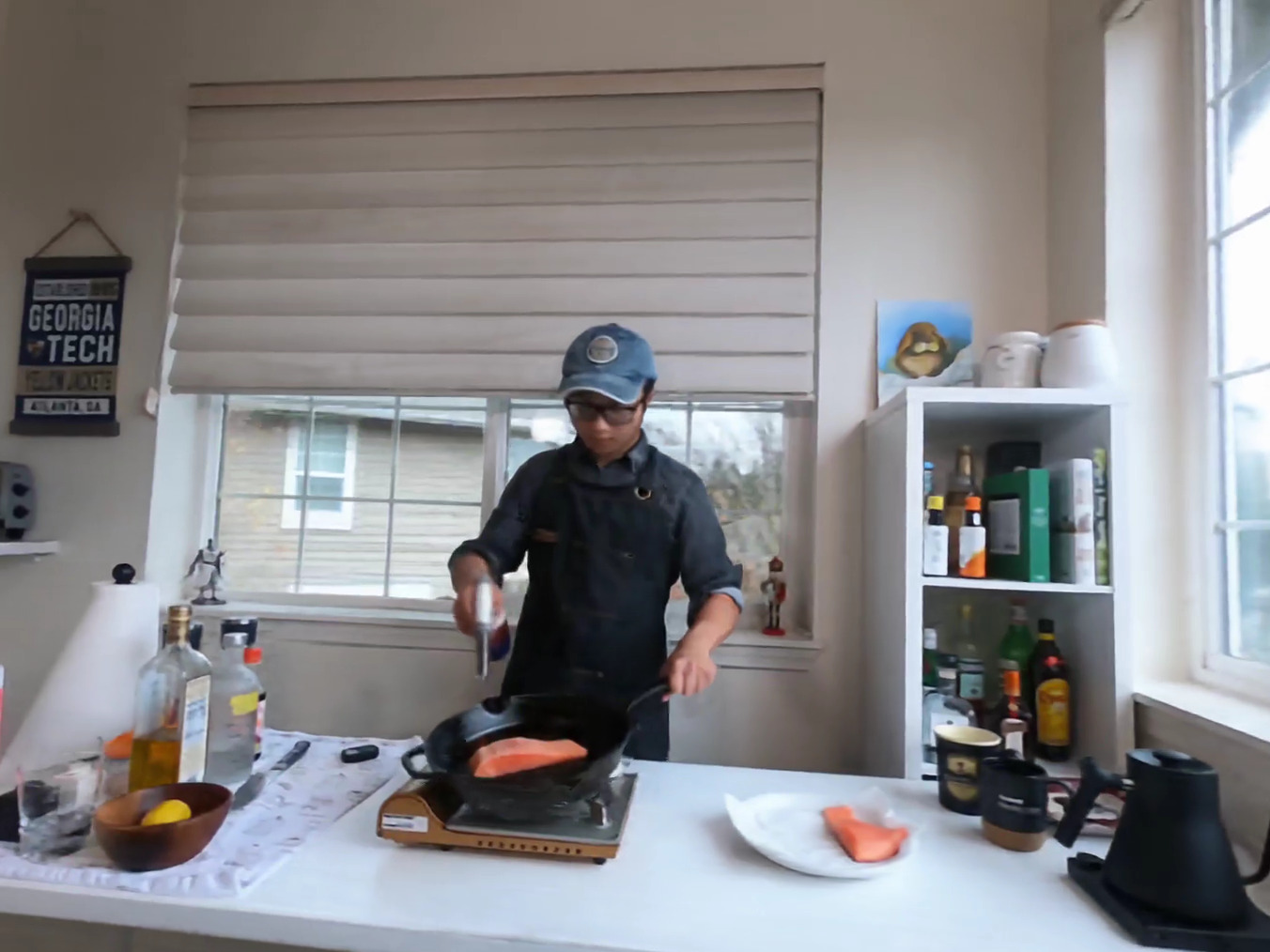} &
    \videozoomed{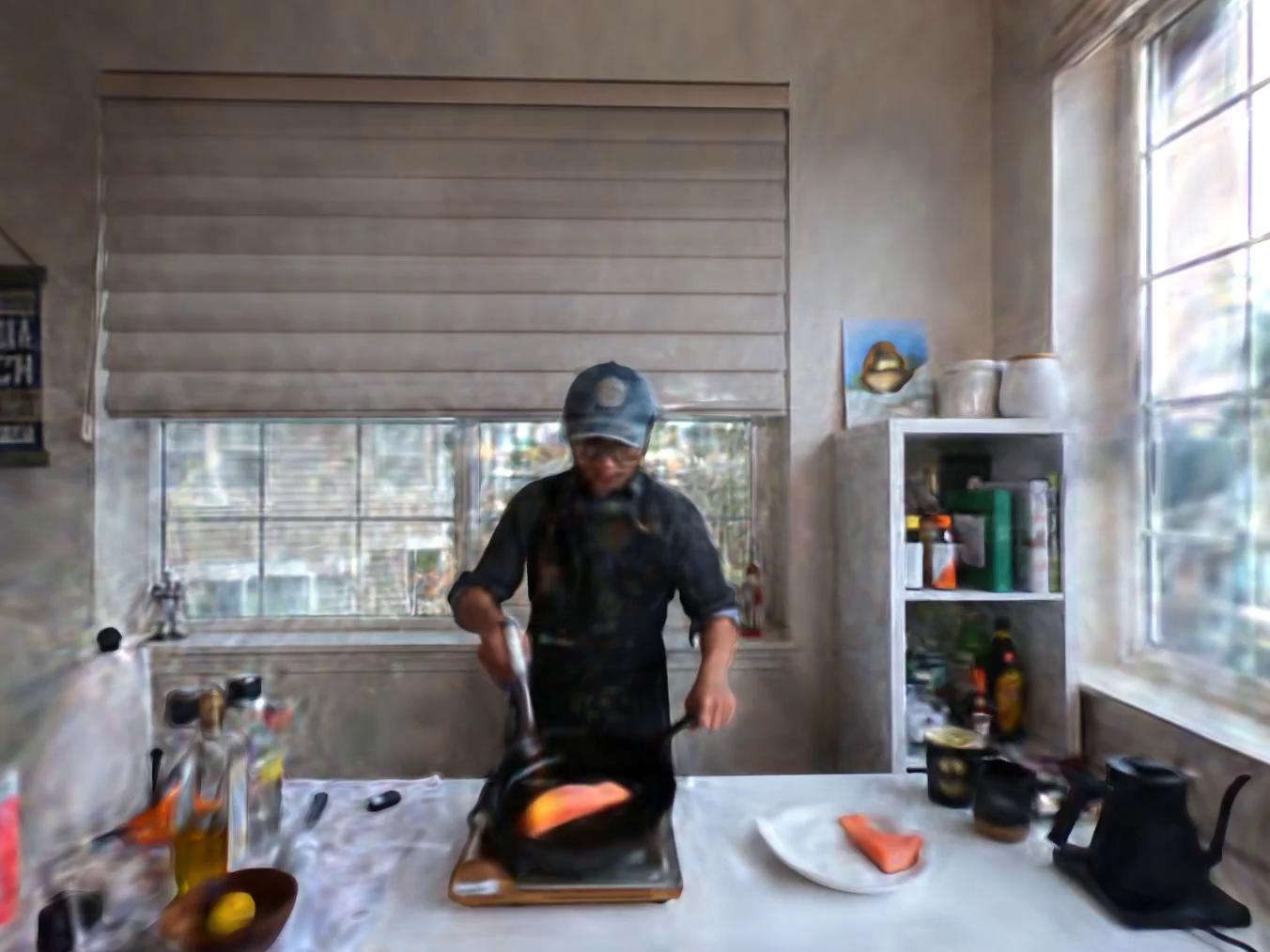} &
    \videozoomed{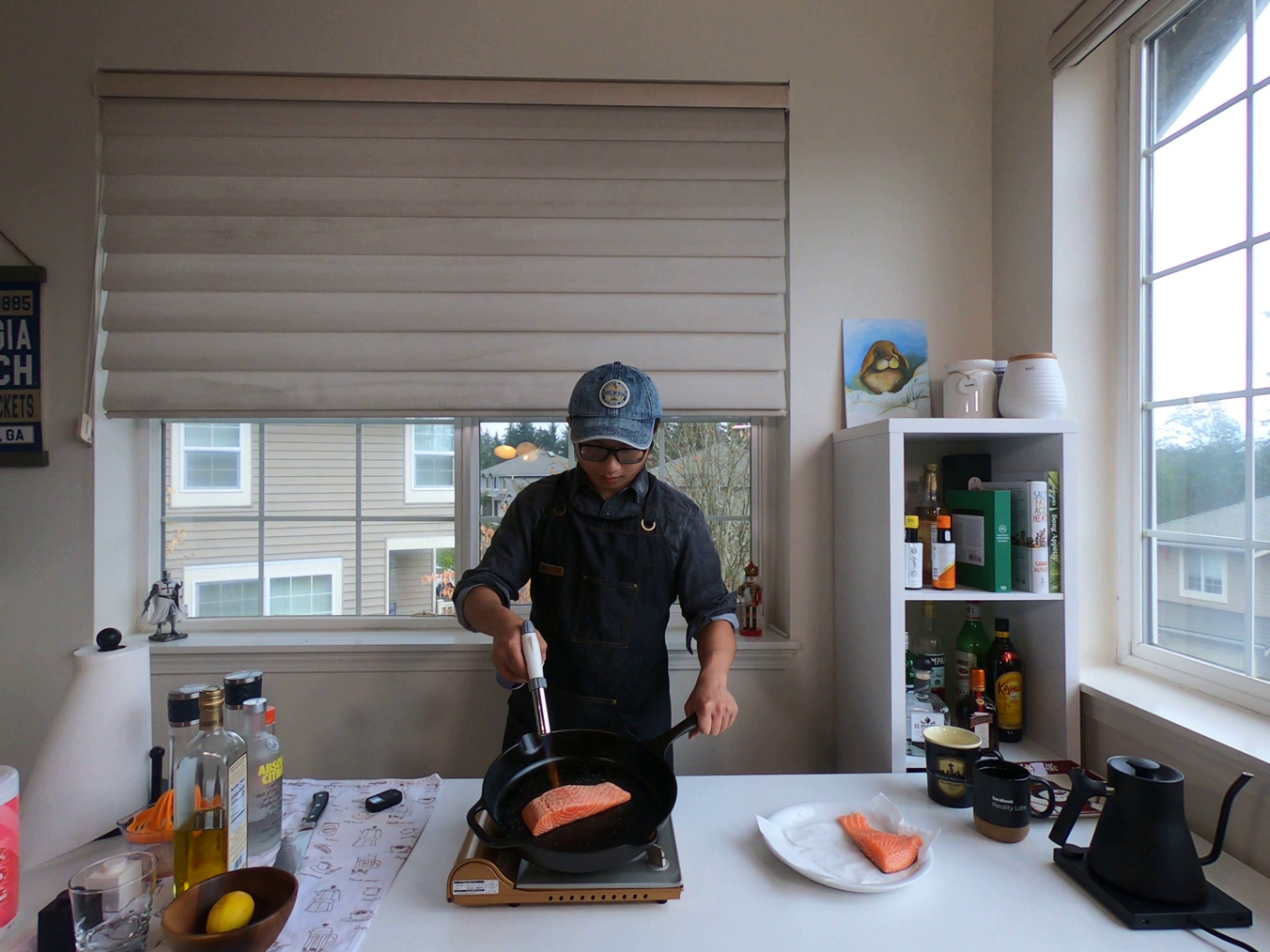} 
    \\ [-10mm]
    \multicolumn{1}{c}{\small Ours-explicit} &
    \multicolumn{1}{c}{\small Ours-hybrid} &
    \multicolumn{1}{c}{\small DyNeRF} &
    \multicolumn{1}{c}{\small MixVoxels} &
    \multicolumn{1}{c}{\small Neural Volumes} &
    \multicolumn{1}{c}{\small Ground truth}
    \end{tabular}
    \caption{\textbf{Qualitative video results.} Our hexplane model rivals the rendering quality of state-of-the-art neural rendering methods. Our renderings were obtained after at most 4 hours of optimization on a single GPU whereas DyNeRF trained for a week on 8 GPUs. MixVoxels frame comes from a slightly different video rendering, and is thus slightly shifted.}
    \label{fig:salmon}
    \vspace{3pt}
\end{figure*}

We first demonstrate our triplane model on the bounded, $360^{\circ}$, synthetic scenes from NeRF \cite{nerf}. We use a model with three symmetric spatial resolutions $N \in \{128, 256, 512\}$ and feature length $M = 32$ at each scale; please see the appendix for ablation studies over these hyperparameters. 
The explicit and hybrid versions of our model perform similarly, within the range of recent results on this benchmark.
% {\color{red}The explicit version of our model matches the prior state-of-the-art in terms of quality metrics, while the hybrid version achieves slightly higher quality metrics}. 
\cref{fig:synthetic-zoom} shows zoomed-in visual results on a small sampling of scenes. 
We also present results of our triplane model on the unbounded, forward-facing, real scenes from LLFF \cite{llff}. 
Our results on this dataset are similar to the synthetic static scenes; both versions of our model match or exceed the prior state-of-the-art, with the hybrid version achieving slightly higher metrics than the fully explicit version. \cref{fig:llff-zoom} shows zoomed-in visual results on a small sampling of scenes.

\subsection{Dynamic scenes}
\label{sec:video}

We evaluate our hexplane model on two dynamic scene datasets: a set of synthetic, bounded, $360^{\circ}$, monocular videos from D-NeRF \cite{dnerf} and a set of real, unbounded, forward-facing, multiview videos from DyNeRF \cite{dynerf}.

The D-NeRF dataset contains eight videos of varying duration, from 50 frames to 200 frames per video. Each timestep has a single training image from a different viewpoint; the camera ``teleports'' between adjacent timestamps~\cite{gao22teleport}. Standardized test views are from novel camera positions at a range of timestamps throughout the video. Both our explicit and hybrid models outperform D-NeRF in both quality metrics and training time, though they do not surpass very recent hybrid methods TiNeuVox \cite{tineuvox} and V4D \cite{v4d}, as shown in \cref{fig:dnerf-zoom}.

The DyNeRF dataset contains six 10-second videos recorded at 30 fps simultaneously by 15-20 cameras from a range of forward-facing view directions; the exact number of cameras varies per scene because a few cameras produced miscalibrated videos. A central camera is reserved for evaluation, and training uses frames from the remaining cameras. Both our methods again produce similar quality metrics to prior state-of-the-art, including recent hybrid method MixVoxels \cite{mixvoxels}, with our hybrid method achieving higher quality metrics. See \cref{fig:salmon} for a visual comparison.

% trim: left bottom right top
\newcommand{\plotcroptrex}[1]{%
  \adjincludegraphics[trim={{0.45\width} {0.3\height} {0.25\width} {0.45\height}}, clip, width=\linewidth]{#1}%
}
\newcommand{\plotcroplego}[1]{%
  \adjincludegraphics[trim={{0.4\width} {0.65\height} {0.3\width} {0.1\height}}, clip, width=\linewidth]{#1}%
}
\begin{figure}
  \centering
  \begin{subfigure}[b]{0.24\linewidth}
    \plotcroptrex{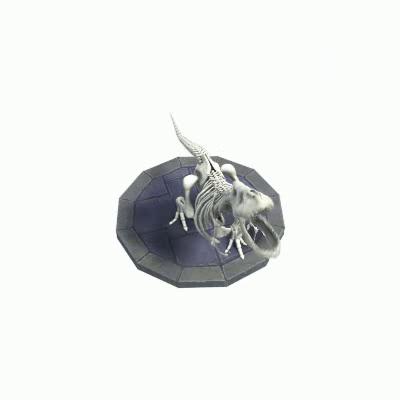}
    \plotcroplego{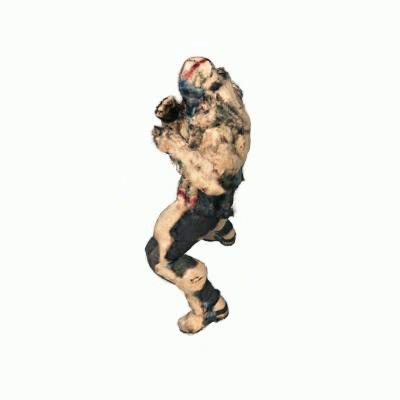}
    \caption{Ours-explicit}
  \end{subfigure}
  \begin{subfigure}[b]{0.24\linewidth}
    \plotcroptrex{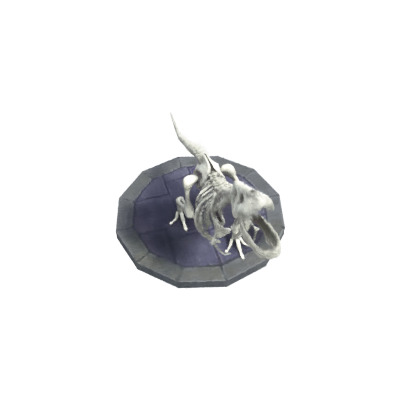}
    \plotcroplego{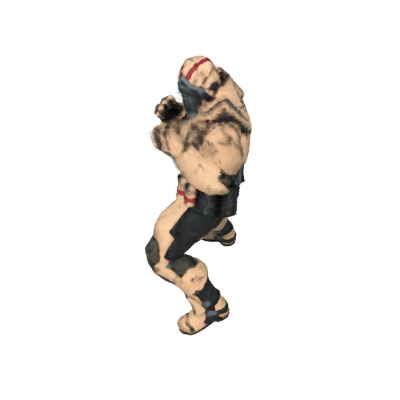}
    \caption{TiNeuVox}
  \end{subfigure}
  \begin{subfigure}[b]{0.24\linewidth}
    \plotcroptrex{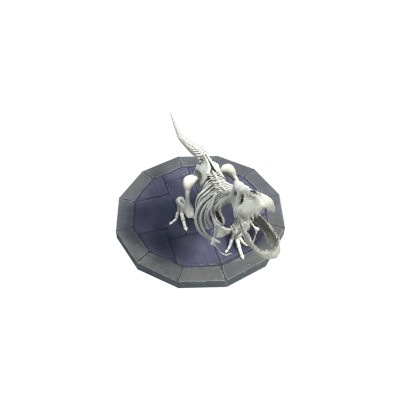}
    \plotcroplego{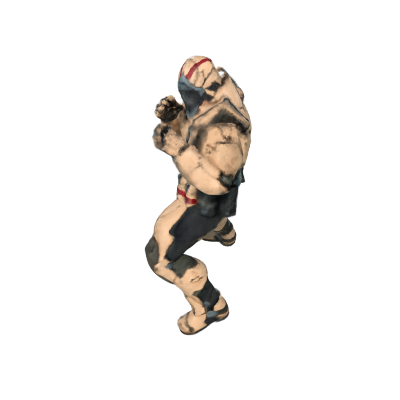}
    \caption{V4D}
  \end{subfigure}
  \begin{subfigure}[b]{0.24\linewidth}
    \plotcroptrex{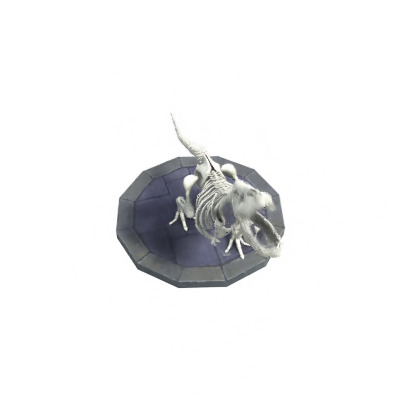}
    \plotcroplego{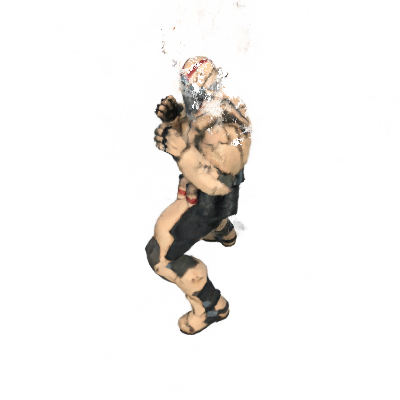}
    \caption{D-NeRF}
  \end{subfigure}
  \caption{\textbf{Zoomed qualitative results on scenes from D-NeRF~\cite{dnerf}.} Visual comparison of \modelname{}, D-NeRF~\cite{dnerf}, TiNeuVox~\cite{tineuvox} and V4D~\cite{v4d}, on \emph{t-rex} (top) and \emph{hook} (bottom).}
  \label{fig:dnerf-zoom}
\end{figure}

\subsubsection{Decomposing time and space}

\begin{figure*}[t]
    \centering
    \vspace{-2mm}
    \includegraphics[width=0.45\linewidth]{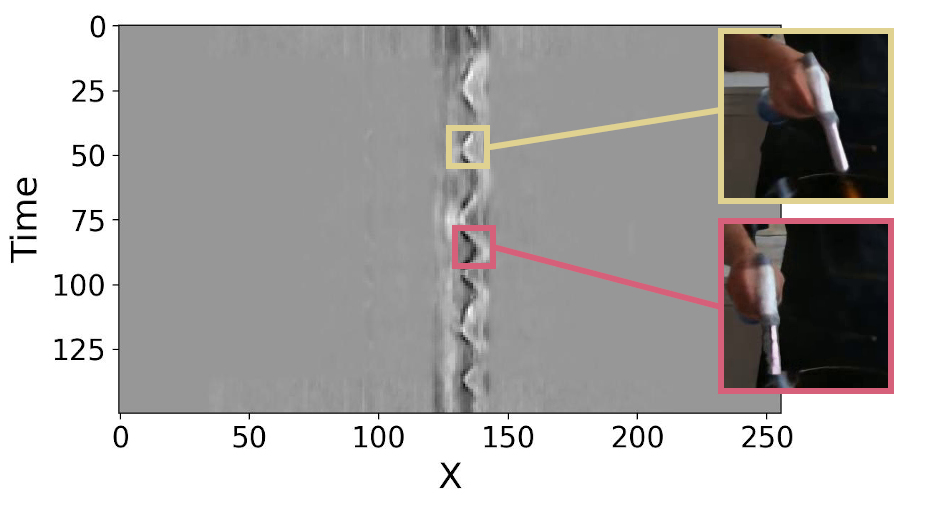}%\vspace{-5mm}  % hides X label from top fig.
    %\hspace*{1pt}
    \includegraphics[width=0.45\linewidth]{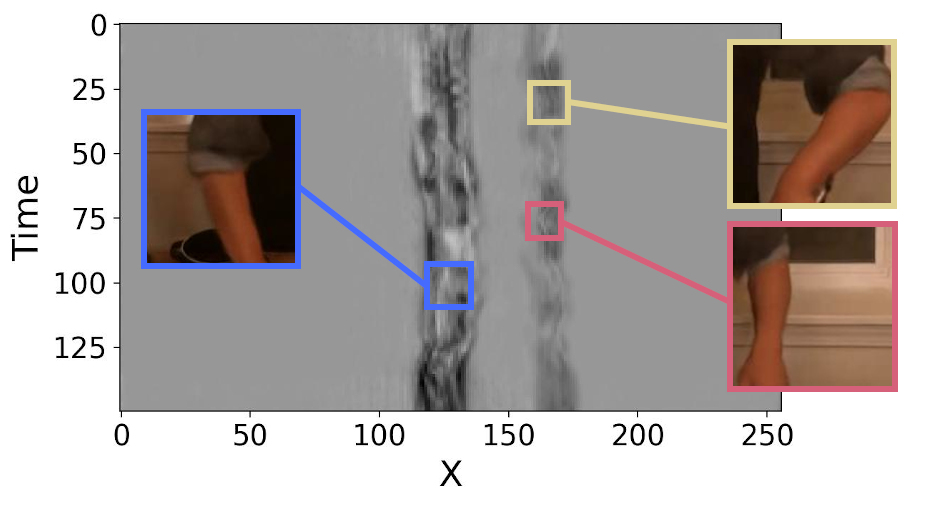}   
    \vspace{-2mm}
    \caption{\textbf{Visualization of a time plane.} The $xt$ plane highlights the dynamic regions in the scene. The wiggly patterns across time correspond to the motion of the person's hands and cooking tools, in the \emph{flame salmon} scene (left) where only one hand moves and the \emph{cut beef} scene (right) where both hands move.
    }
    \label{fig:timeplanes}
\end{figure*}   

One neat consequence of our planar decomposition of time and space is that it naturally disentangles dynamic and static portions of the scene. The static-only part of the scene can be obtained by setting the three time planes to one (the multiplicative identity). Subtracting the static-only rendered image from the full rendering (i.e. with the time plane parameters not set to \num{1}), we can reveal the dynamic part of the scene.
\cref{fig:video-decomposition} shows this decomposition of time and space. This natural volumetric disentanglement of a scene into static and dynamic regions may enable many applications across augmented and virtual reality~\cite{volumentricdisentanglement}.

We can also visualize the time planes to better understand  where motion occurs in a video. \cref{fig:timeplanes} shows the averaged features learned by the $xt$ plane in our model for the \emph{flame salmon} and \emph{cut beef} DyNeRF videos, in which we can identify the motions of the hands in both space and time. 
The $xt$ plane learns to be sparse, with most entries equal to the multiplicative identity, due to a combination of our sparse transients prior and the true sparsity of motion in the video. For example, in the left side of \cref{fig:timeplanes} one of the cook's arms contains most of the motion, while in the right side both arms move. 
Having access to such an explicit representation of time allows us to add time-specific priors.

\newcommand{\plotcropsalmon}[2]{%
    \begin{overpic}[tics=10, trim=300 60 350 280, clip=true, width=0.3\linewidth]{#1}
        \put (70,80) {\makebox[0pt]{\centering \color{black} {#2}}}
    \end{overpic}
}
\newcommand{\plotdnerf}[2]{%
    \begin{overpic}[tics=10, trim=40 40 40 40, clip=true, width=0.3\linewidth]{#1}
        \put (20,80) {\makebox[0pt]{\centering \color{black} {#2}}}
    \end{overpic}
}
\begin{figure}[t]
    \begin{minipage}[]{1.0\linewidth}
    \centering
    \begin{tabular}{c@{}c@{}c@{}}
        \plotcropsalmon{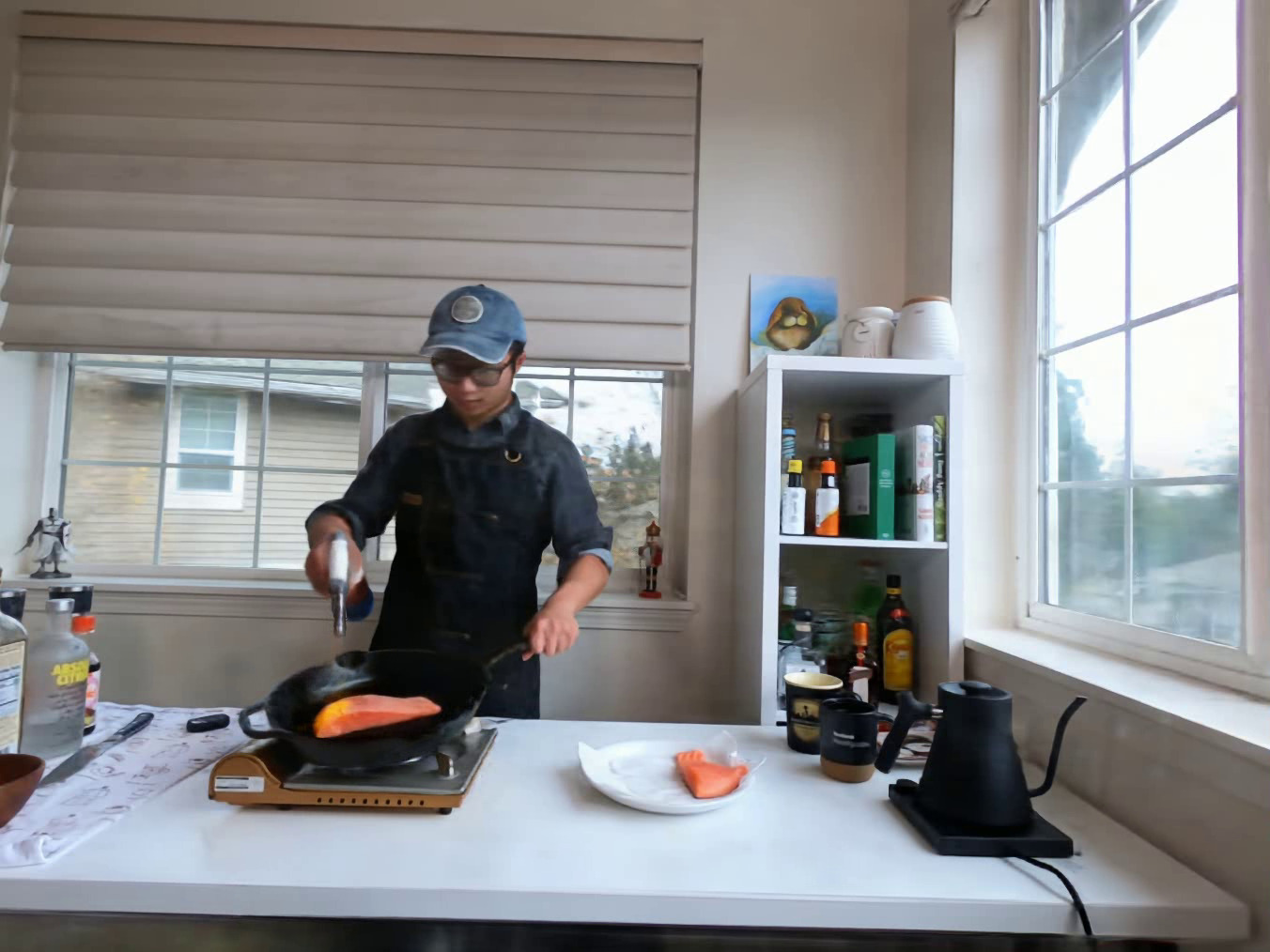}{Full} &
        \plotcropsalmon{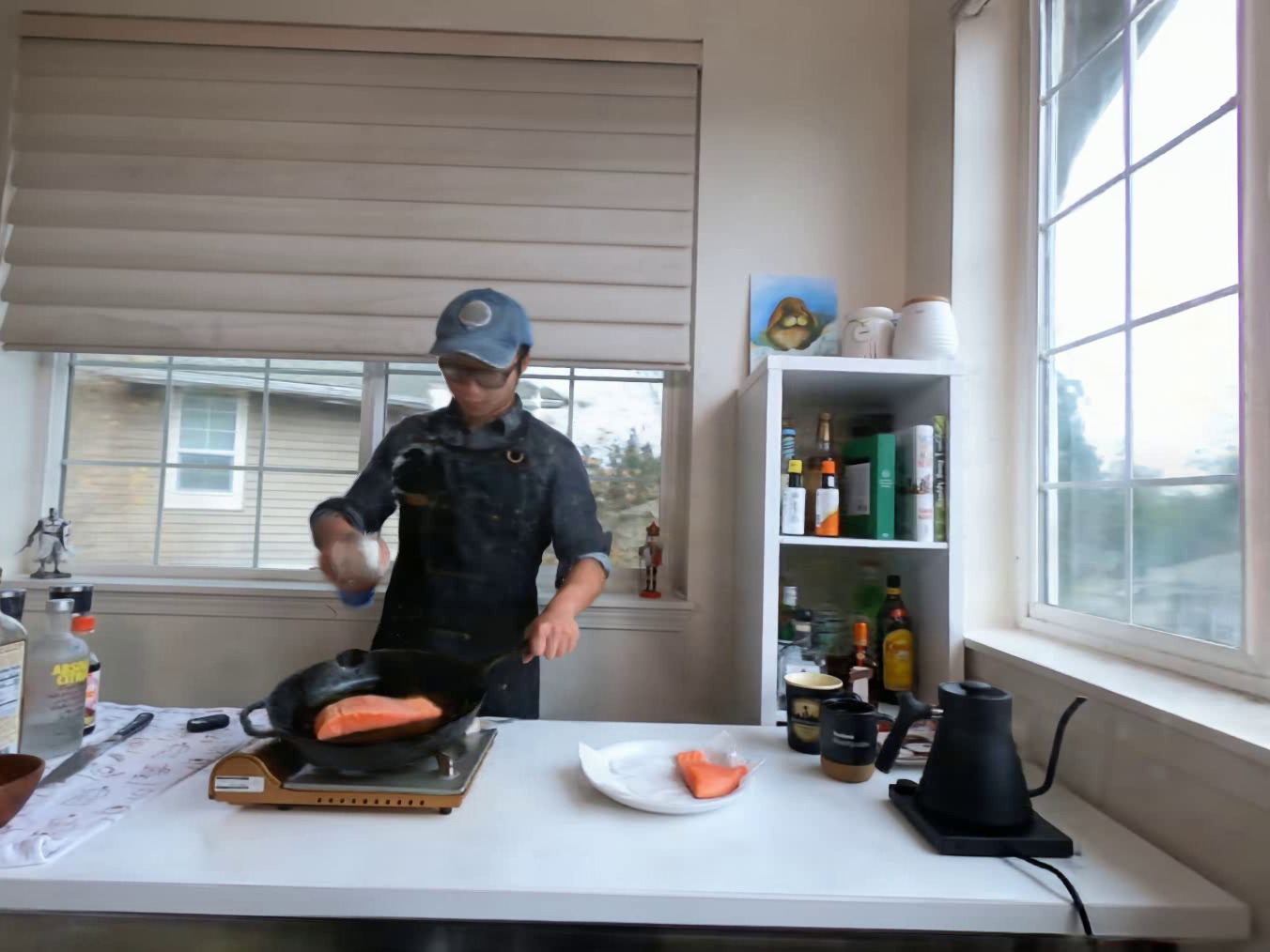}{Space} &
        \plotcropsalmon{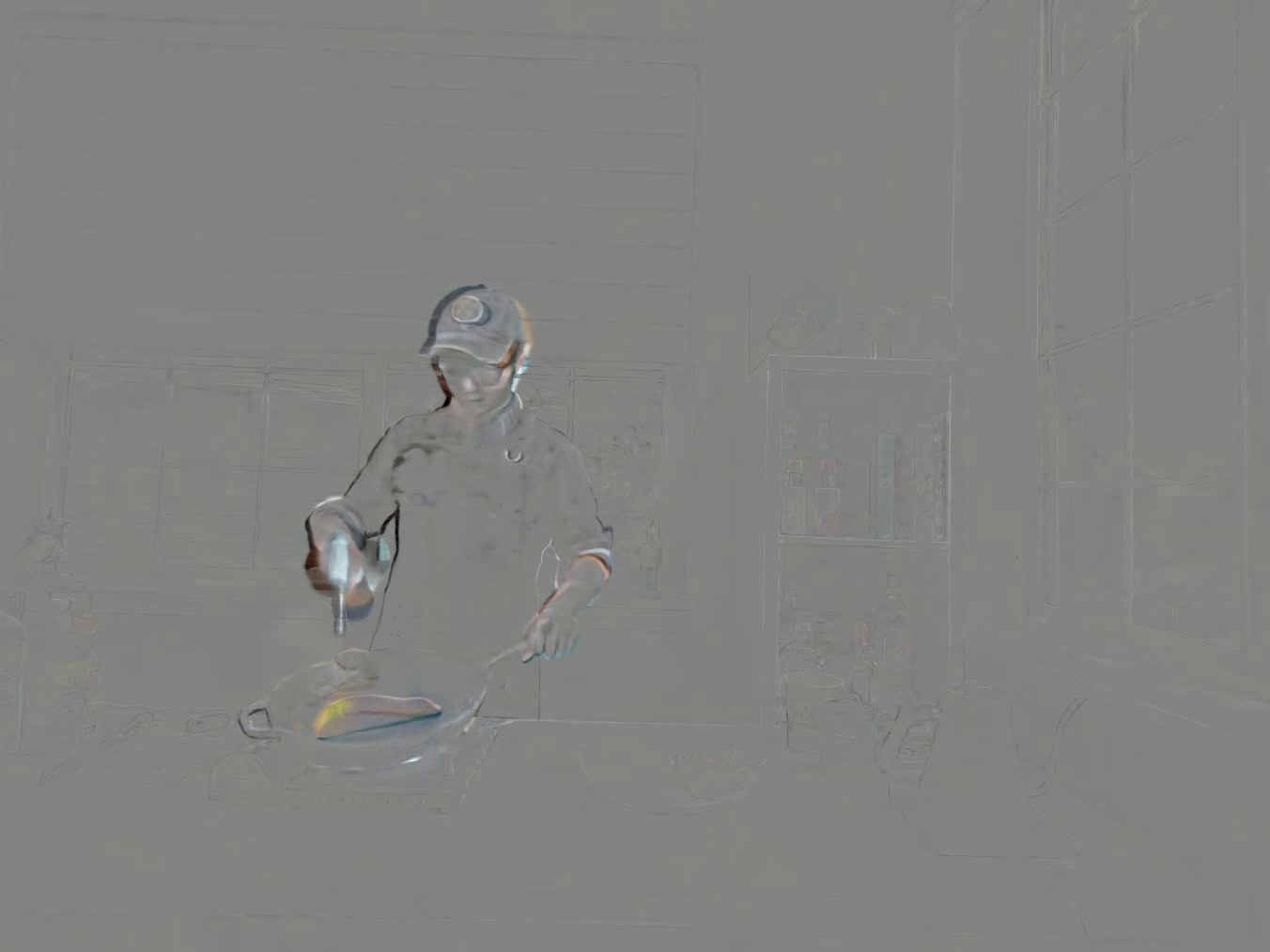}{Time} \\
        \plotdnerf{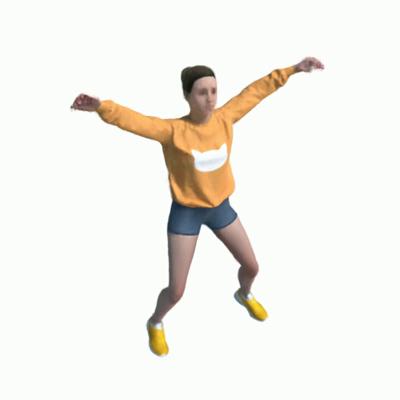}{Full} &
        \plotdnerf{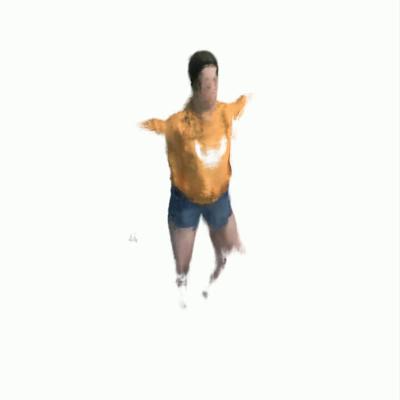}{Space} &
        \plotdnerf{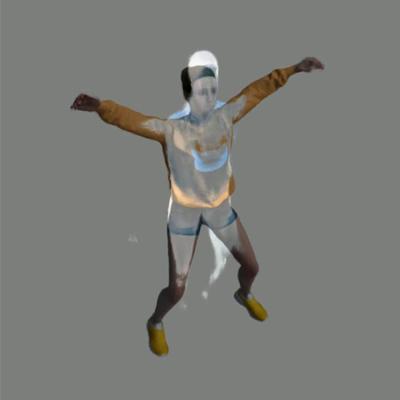}{Time} \\
    \end{tabular}
    \end{minipage}
    \caption{\textbf{Decomposition of space and time.} \Modelname{} (left) naturally decomposes a 3D video into static and dynamic components. We render the static part (middle) by setting the time planes to the identity, and the remainder (right) is the dynamic part. Top shows the \emph{flame salmon} multiview video \cite{dynerf} and bottom shows the \emph{jumping jacks} monocular video \cite{dnerf}.}\label{fig:video-decomposition}
\end{figure}

\subsection{Variable appearance}
\label{sec:appearance}

Our variable appearance experiments use the Phototourism dataset \cite{phototourism}, which includes photos of well-known landmarks taken by tourists with arbitrary view directions, lighting conditions, and transient occluders, mostly other tourists. Our experimental conditions parallel those of NeRF-W \cite{martinbrualla2020nerfw}: we train on more than a thousand tourist photographs and test on a standard set that is free of transient occluders. Like NeRF-W, we evaluate on test images by optimizing our per-image appearance feature on the left half of the image and computing metrics on the right half. Visual comparison to prior work is shown in the appendix.%visualized in \cref{fig:appearance}.

Also similar to NeRF-W~\cite{martinbrualla2020nerfw, Piotr2017latentspaceoptim}, we can interpolate in the appearance code space. Since only the color decoder (and not the density decoder) takes the appearance code as input, our approach is guaranteed not to change the geometry, regardless of whether we use our explicit or our hybrid model. %In contrast to NeRF-W, which uses a 4-layer MLP with 128 hidden units and an appearance code of 48-dimensional appearance code, 
\cref{fig:interpolation} shows that our planar decomposition with a 32-dimensional appearance code is sufficient to accurately capture global appearance changes in the scene.

\newcommand{\figinterpheight}{2.475cm}
\begin{figure}
    \centering
    \begin{tabular}{c@{\hskip 1mm}c@{}@{\hskip 1mm}c@{}}
    \includegraphics[height=\figinterpheight]{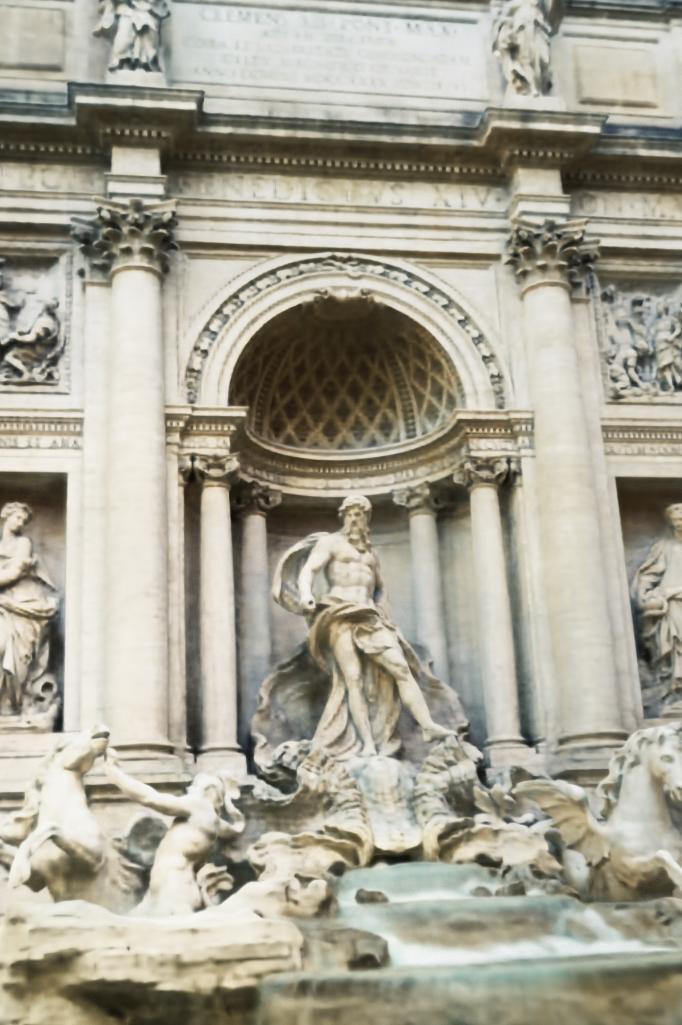} &
    \includegraphics[height=\figinterpheight]{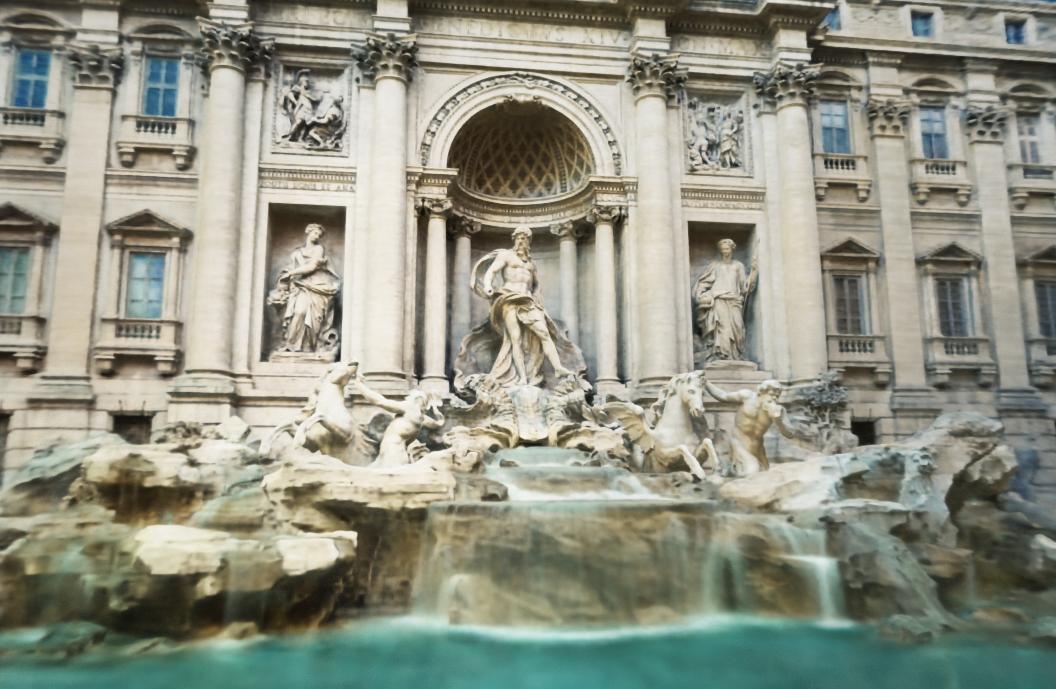} &
    \includegraphics[height=\figinterpheight]{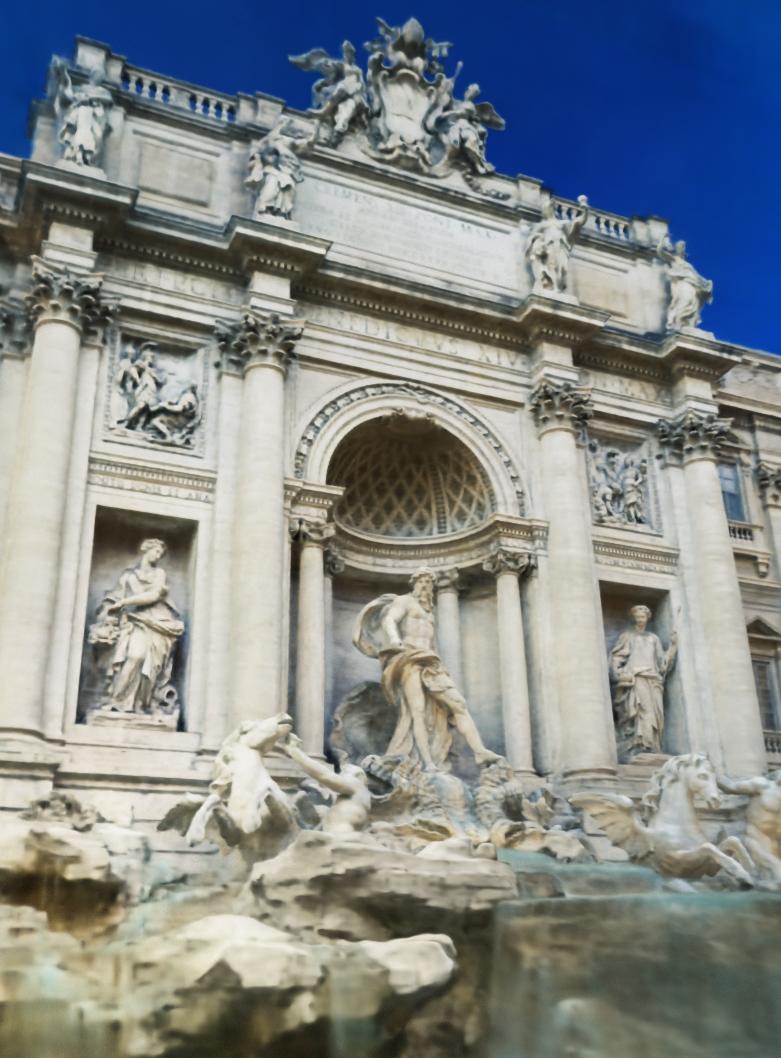} \\
    \includegraphics[height=\figinterpheight]{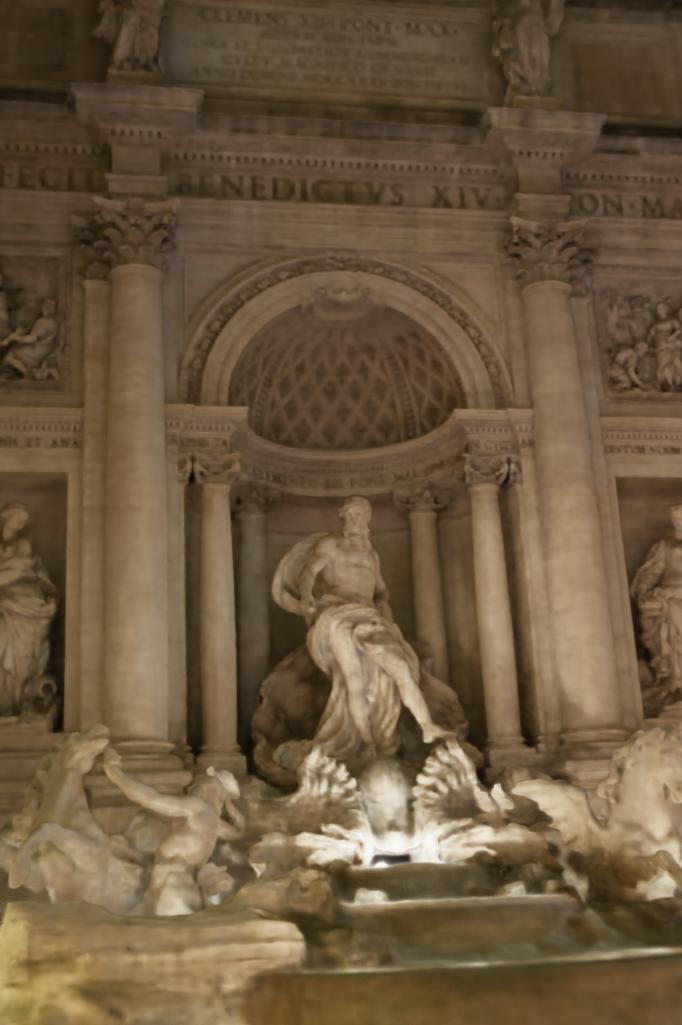} &
    \includegraphics[height=\figinterpheight]{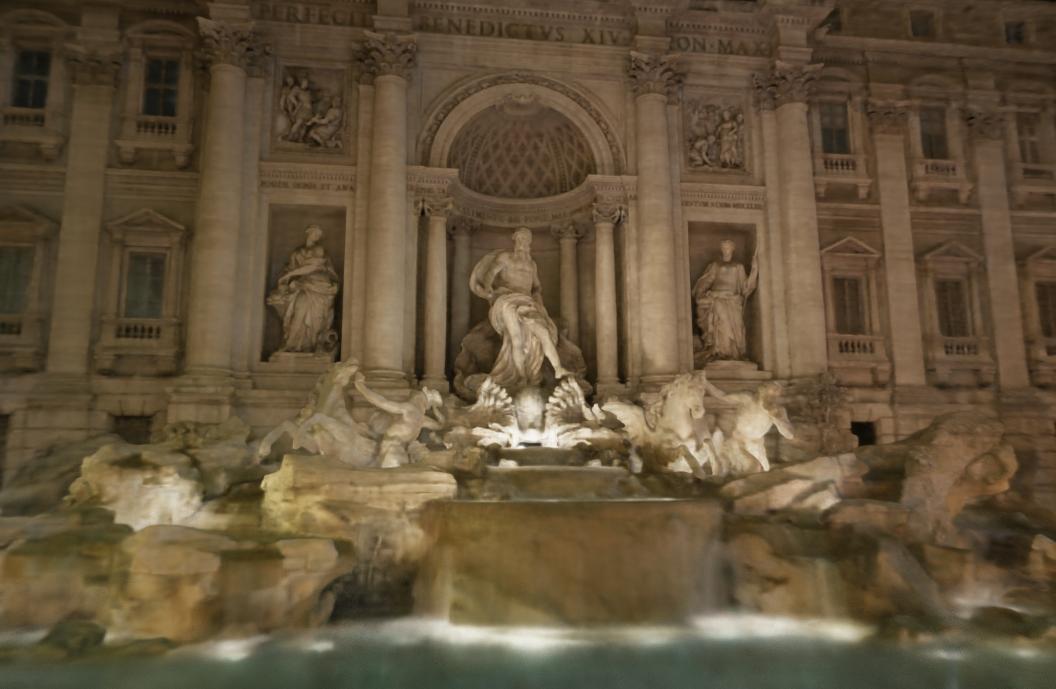} &
    \includegraphics[height=\figinterpheight]{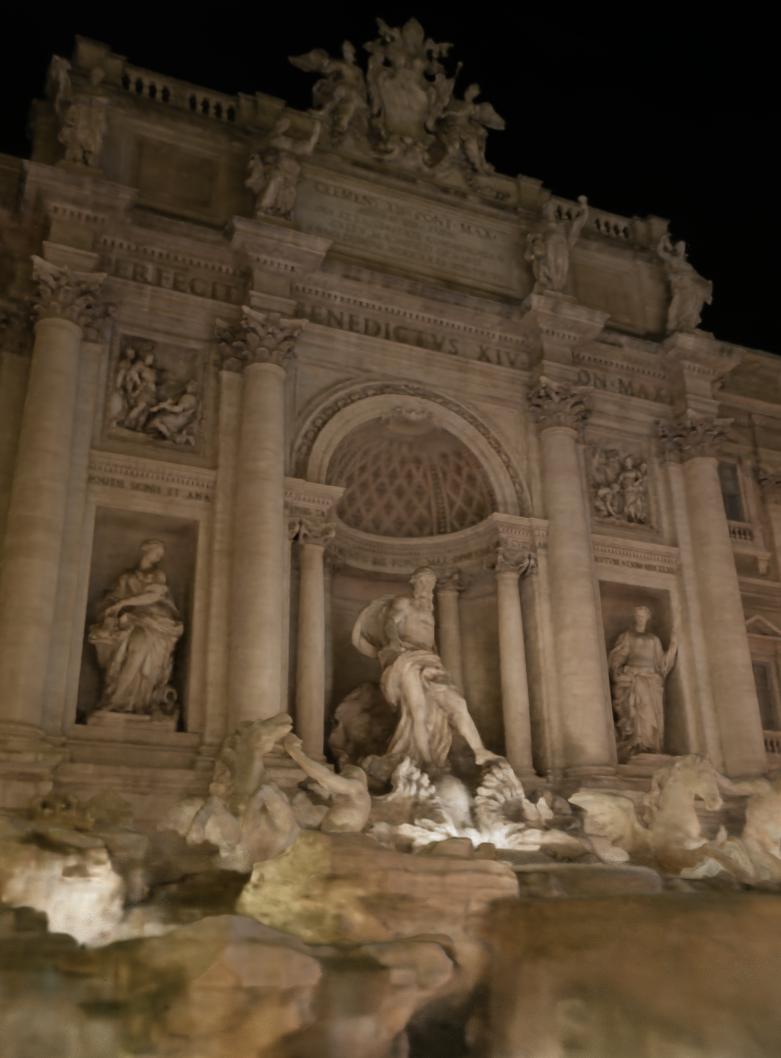}
    \end{tabular} 
    \caption{\textbf{Appearance interpolation}. Like NeRF-W~\cite{martinbrualla2020nerfw}, we can interpolate our appearance code to alter the visual appearance of landmarks. We show three test views from the \emph{Trevi fountain} with appearance codes corresponding to day and night.}
    \label{fig:interpolation}
\end{figure}

\section{Conclusions}

We introduced a simple yet versatile method to decompose a $d$-dimensional space into $\binom{d}{2}$ planes, which can be optimized directly from indirect measurements and scales gracefully in model size and optimization time with increasing dimension, without any custom CUDA kernels. 
We demonstrated that the proposed \modelname{} decomposition applies naturally to reconstruction of static 3D scenes as well as dynamic 4D videos, and with the addition of a global appearance code can also extend to the more challenging task of unconstrained scene reconstruction. \Modelname{} is the first explicit, simple model to demonstrate competitive performance across such varied tasks.

\paragraph{Acknowledgments.} \looseness=-1 Many thanks to Matthew Tancik, Ruilong Li, and other members of KAIR for helpful discussion and pointers. We also thank the DyNeRF authors for their response to our questions about their method.

\newpage

%%%%%%%%% REFERENCES
{\small
\bibliographystyle{ieee_fullname}
\bibliography{egbib}
}
\clearpage
\section{Appendix}

\subsection{Volumetric rendering}

We use the same volume rendering formula as NeRF \cite{nerf}, originally from \cite{max1995}, where the color of a pixel is represented as a sum over samples taken along the corresponding ray through the volume:

\begin{align}
\label{eq:max}
    \sum_{i=1}^N \exp \left(-\sum_{j=1}^{i-1} \sigma_j \delta_j \right) \big(1 - \exp(-\sigma_i \delta_i)\big)\textbf{c}_i 
\end{align}
where the first $\exp$ represents ray transmission to sample~$i$, $1 - \exp(-\sigma_i \delta_i)$ is the absorption by sample $i$,  $\sigma_i$ is the (post-activation) density of sample $i$, and $\textbf{c}_i$ is the color of sample $i$, with distance $\delta_i$ to the next sample. 

\subsection{Per-scene results}

\newcommand\trevipic[1]{
    \raisebox{-0.5\height}{
    \begin{tikzpicture}[
    zoomboxarray,
    zoomboxes below,
    connect zoomboxes,
    zoombox paths/.append style={ultra thick}]
        \node[image node]{\includegraphics[width=0.148\textwidth]{figures/appearance/trevi/#1.jpg}};
        \zoombox[magnification=5,color code=col1]{0.681,0.595}
        \zoombox[magnification=6,color code=col2]{0.715,0.495}  
    \end{tikzpicture}}
}
\newcommand\brandenburgpic[1]{
    \raisebox{-0.5\height}{
    \begin{tikzpicture}[
    zoomboxarray,
    zoomboxes below,
    connect zoomboxes,
    zoombox paths/.append style={ultra thick}]
        \node[image node]{\includegraphics[width=0.154\textwidth]{figures/appearance/brandenburg/#1.jpg}};
        \zoombox[magnification=4,color code=col1]{0.48,0.86}
        \zoombox[magnification=4,color code=col2]{0.54,0.55}
    \end{tikzpicture}}
}
\begin{figure*}[t]
    \centering 
    \def\arraystretch{1}
    \begin{tabular}{c@{\hskip 2mm}c@{}c@{}c@{}c@{}c@{}c@{}}
        \rotatebox[origin=c]{90}{\hspace{10mm} Trevi Fountain} &
        \trevipic{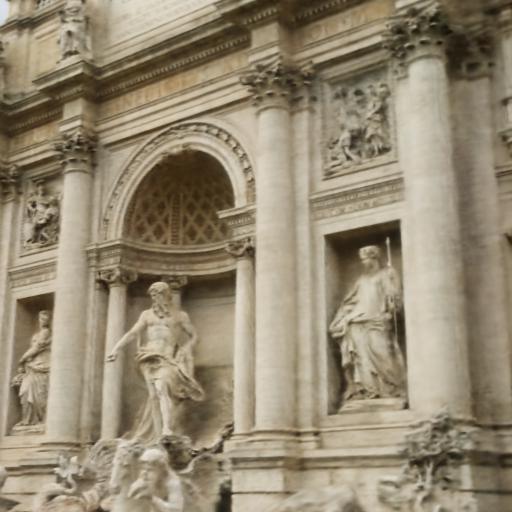} &
        \trevipic{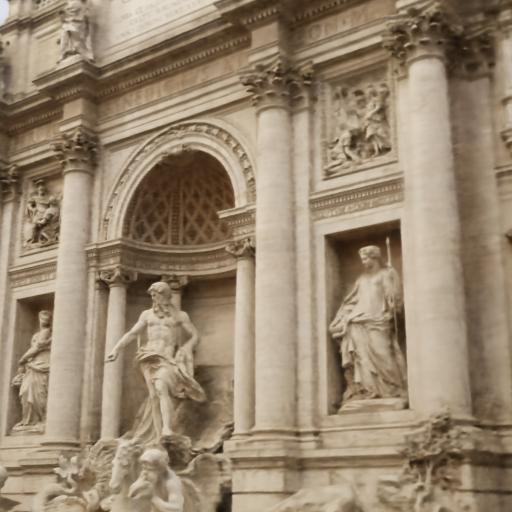} &
        \trevipic{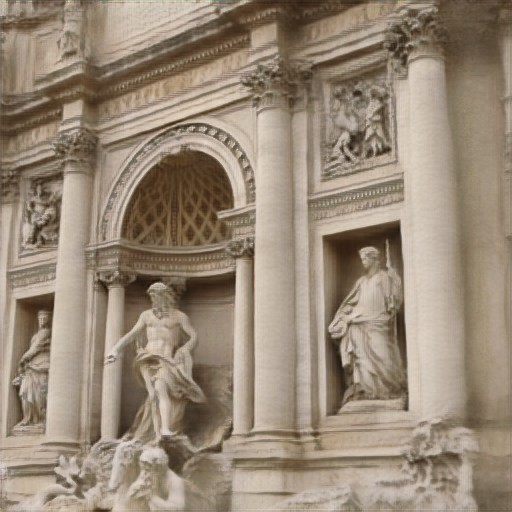} &
        \trevipic{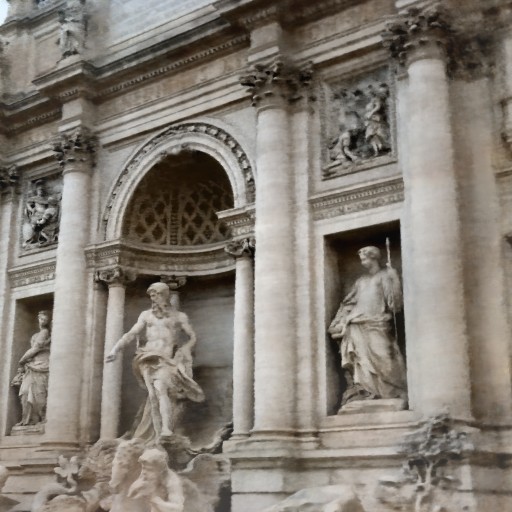} &
        \trevipic{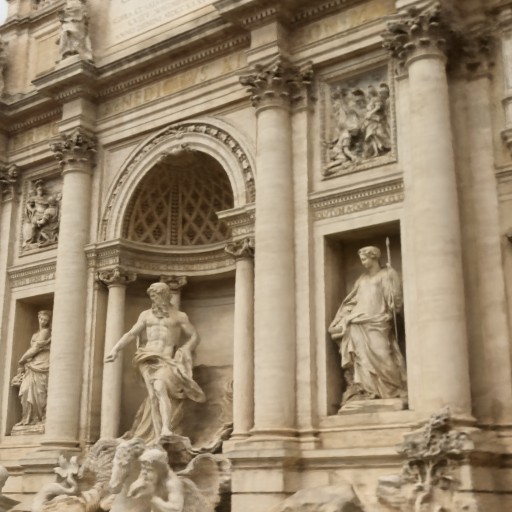} &
        \trevipic{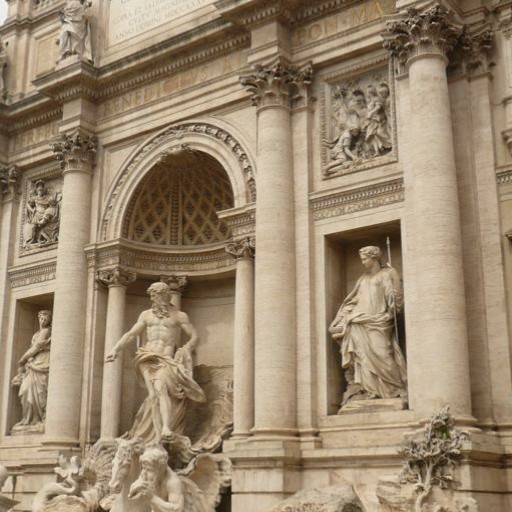}
    \\ [-11mm]
    &
    \multicolumn{1}{c}{\small Ours-explicit} &
    \multicolumn{1}{c}{\small Ours-hybrid} &
    \multicolumn{1}{c}{\small NRW \cite{nrw}} &
    \multicolumn{1}{c}{\small NeRF} &
    \multicolumn{1}{c}{\small NeRF-W} &
    \multicolumn{1}{c}{\small Ground truth}
    \\
    \end{tabular}
    \caption{
        \textbf{Qualitative results from Phototourism dataset}. We compare our model with strong baselines. Our method captures the geometry and appearance of the scene, but produces slightly lower resolution results than NeRF-W. Note that our model optimizes in just 35 minutes on a single GPU compared to NeRF-W, which takes 2 days on 8 GPUs. 
    } 
    \label{fig:appearance}
\end{figure*}

\cref{fig:appearance} provides a qualitative comparison of methods for the Phototourism dataset, on the \emph{Trevi fountain} scene. We also provide quantitative metrics for each of the three tasks we study, for each scene individually. \cref{tab:fullsynthetic} reports metrics on the static synthetic scenes, \cref{tab:full_llff} reports metrics on the static real forward-facing scenes, \cref{tab:fulldnerf} reports metrics on the dynamic synthetic monocular ``teleporting camera'' scenes, \cref{tab:fullvideo} reports metrics on the dynamic real forward-facing multiview scenes, and \cref{tab:fullphototourism} reports metrics on the Phototourism scenes.

\subsection{Ablation studies}

\paragraph{Multiscale.}
In \cref{tab:multiscale}, we ablate our model on the static \emph{Lego} scene \cite{nerf} with respect to our multiscale planes, to assess the value of including copies of our model at different scales. 

\renewcommand{\tabcolsep}{6pt}
\begin{table}[ht]
  \centering
  \begin{tabular}{rlccr}
    \multicolumn{5}{c}{} \\
    \toprule
     Scales & & Explicit & Hybrid  &  \\ %& GPU hrs $\downarrow$\\ 
     (32 Feat. Each) & & PSNR $\uparrow$ & PSNR $\uparrow$ & \# params $\downarrow$ \\
    \cmidrule(){1-1} \cmidrule(){3-5}
    $64,128,256,512$ &&	35.26 & 35.79 & 34M \\
    $128,256,512$ && 35.29 & 35.75 & 33M \\
    $256,512$ && 34.52 & 35.37 & 32M \\
    $512$ && 32.93 & 33.60 & 25M \\
    $64,128,256$\hspace{0.67cm} && 34.26 & 35.07 & 8M	\\
    % $64, 128, 256, 512$ && 36.36 & 36.97 & 34M \\
    % $128, 256, 512$ &&	36.44 &	37.23 &	34M \\
    % $256, 512$ &&	36.40 &	37.19 &	32M \\
    % $512$ &&	35.65 &	36.92 &	26M \\
    % $64, 128, 256$\hspace{0.67cm} &&	33.79 &	35.22 &	9M \\
    \bottomrule
    \\
    \toprule
     Scales & & Explicit & Hybrid  &  \\ %& GPU hrs $\downarrow$\\ 
     (96 Feat. Total) & & PSNR $\uparrow$ & PSNR $\uparrow$ & \# params $\downarrow$ \\
    \cmidrule(){1-1} \cmidrule(){3-5} 
    $64,128,256,512$ && 35.16 & 35.67 & 25M \\
    $128,256,512$ && 35.29 & 35.75 & 33M \\
    $256,512$ && 34.50 & 35.16 & 47M \\
    $512$ && 33.12 & 34.09 & 76M \\
    $64,128,256$\hspace{0.67cm} && 34.26 & 35.07 & 8M \\
    \bottomrule
  \end{tabular}
  \caption{\textbf{Ablation study over scales.} 
  Including even a single lower scale improves performance, for both our explicit and hybrid models, even when holding the total feature dimension constant. Using lower scales only (excluding resolution $512^3$) substantially reduces model size and yields quality much better than using high resolution alone, though slightly worse than including both low and high resolutions. This experiment uses the static \emph{Lego} scene; in the top table each scale is allocated 32 features and in the bottom table a total of 96 features are allocated evenly among all scales.}
  \label{tab:multiscale}
\end{table}

\paragraph{Feature length.}
In \cref{tab:featurelen}, we ablate our model on the static \emph{Lego} scene with respect to the feature dimension $M$ learned at each scale.

\renewcommand{\tabcolsep}{6pt}
\begin{table}[ht]
  \centering
  \begin{tabular}{rlccr}
    \multicolumn{5}{c}{} \\
    \toprule
     Feature Length & & Explicit & Hybrid  &  \\ %& GPU hrs $\downarrow$\\ 
    ($M$) & & PSNR $\uparrow$ & PSNR $\uparrow$ & \# params $\downarrow$ \\
    \cmidrule(){1-1} \cmidrule(){3-5} 
    2 && 30.66 & 32.05 & 2M \\
    4 && 32.27 & 34.18 & 4M \\
    8 && 33.80 & 35.12 & 8M \\
    16 && 34.80 & 35.44 & 17M \\
    32 && 35.29 & 35.75 & 33M \\
    64 && 35.38 & 35.88 & 66M \\
    128 && 35.45 & 35.99 & 132M \\
    % 4 &&	32.82 &	34.88 &	5M \\
    % 8 &&	34.65 &	35.96 &	9M \\
    % 16 &&	35.57 &	36.78 &	17M \\
    % 32 &&	36.36 &	36.97 &	34M \\
    % 64 &&	36.93 &	37.66 &	67M \\
    \bottomrule
  \end{tabular}
  \caption{\textbf{Ablation study over feature length $M$.}  Increasing the feature length $M$ learned at each scale consistently improves quality for both our models, with a corresponding linear increase in model size and optimization time. Our experiments in the main text use a mixture of $M=16$ and $M=32$; for specific applications it may be beneficial to vary $M$ along this tradeoff between quality and model size. This experiment uses the static \emph{Lego} scene with 3 scales: 128, 256, and 512.}
  \label{tab:featurelen}
\end{table}

\paragraph{Time smoothness regularizer.}
\label{sec:ablation_time_reg}
\cref{sec:regularization} describes our temporal smoothness regularizer based on penalizing the norm of the second derivative over the time dimension, to encourage smooth motion and discourage acceleration. \cref{tab:smoothness} illustrates an ablation study of this regularizer on the \emph{Jumping Jacks} scene from D-NeRF \cite{dnerf}.

\renewcommand{\tabcolsep}{6pt}
\begin{table}[ht]
  \centering
  \begin{tabular}{rlcc}
    \multicolumn{4}{c}{} \\
    \toprule
     Time Smoothness & & Explicit   & Hybrid    \\ 
     Weight ($\lambda$) && PSNR $\uparrow$ & PSNR $\uparrow$ \\
    \cmidrule(){1-1} \cmidrule(){3-4} 
    % using half resolution
    0.0\textcolor{white}{00} && 30.45 &
30.86 \\
    0.001 && 31.61 &
32.23 \\
    0.01\textcolor{white}{0} && 32.00 &
32.64 \\ 
    0.1\textcolor{white}{00} && 31.96 &
32.58 \\
    1.0\textcolor{white}{00} && 31.36 &
32.22 \\
    10.0\textcolor{white}{00} && 30.45 &
31.63 \\
    % using full resolution
    % 0.0\textcolor{white}{00} && 29.69 &	29.86 \\
    % 0.001 && 30.78 & 	31.24 \\
    % 0.01\textcolor{white}{0} && 31.17 &	31.56 \\ 
    % 0.1\textcolor{white}{00} && 31.10	& 31.50 \\
    % 1.0\textcolor{white}{00} && 30.67	& 31.22 \\
    % 10.0\textcolor{white}{00} && 29.72	& 30.57 \\
    \bottomrule
  \end{tabular}
  \caption{\textbf{Ablation study over temporal smoothness regularization.} For both models, a temporal smoothness weight of 0.01 is best, with PSNR degrading gradually with over- or under-regularization. This experiment uses the \emph{Jumping Jacks} scene with 4 scales: 64, 128, 256, and 512, and 32 features per scale.}
  \label{tab:smoothness}
\end{table}

\section{Model hyperparameters}

Our full hyperparameter settings are available in the config files in our released code, at \url{https://github.com/sarafridov/K-Planes}.

\renewcommand{\tabcolsep}{6pt}
\begin{table*}
  \centering
  \begin{tabular}{llcccccccclc}
    \multicolumn{12}{c}{PSNR $\uparrow$} \\
    \toprule
    && Chair & Drums & Ficus & Hotdog & Lego & Materials & Mic & Ship && Mean \\
    \cmidrule(){1-1} \cmidrule(){3-10} \cmidrule(){12-12}
    Ours-explicit && 34.82 &	25.72 &	31.2 &	36.65 &	35.29 &	29.49 &	34.00 &	30.51 &&	32.21 \\
    Ours-hybrid && 34.99 &	25.66 &	31.41 &	36.78 &	35.75 &	29.48 &	34.05 &	30.74 &&	32.36 \\
    % Ours-explicit                          && 34.15 & 27.99 & 31.71 & 37.81 & 36.40 & 32.11 & 34.14 & 30.71 && 33.13 \\
    % Ours-hybrid                            && 34.74 & 27.84 & 32.40 & 38.17 & 37.21 & 32.66 & 34.76 & 31.19 && 33.62 \\
    INGP~\cite{ingp}                       && 35.00 & 26.02 & 33.51 & 37.40 & 36.39 & 29.78 & 36.22 & 31.10 && 33.18 \\
    TensoRF~\cite{tensorf}                 && 35.76 & 26.01 & 33.99 & 37.41 & 36.46 & 30.12 & 34.61 & 30.77 && 33.14 \\
    Plenoxels~\cite{plenoxels}             && 33.98 & 25.35 & 31.83 & 36.43 & 34.10 & 29.14 & 33.26 & 29.62 && 31.71 \\
    JAXNeRF~\cite{jaxnerf2020github, nerf} && 34.20 & 25.27 & 31.15 & 36.81 & 34.02 & 30.30 & 33.72 & 29.33 && 31.85 \\
    \bottomrule
    \multicolumn{12}{c}{SSIM $\uparrow$} \\
    \toprule
    && Chair & Drums & Ficus & Hotdog & Lego & Materials & Mic & Ship && Mean \\ 
    \cmidrule(){1-1} \cmidrule(){3-10} \cmidrule(){12-12}
    Ours-explicit && 0.981 &	0.937 &	0.975 &	0.982 &	0.978 &	0.949 &	0.988 &	0.892 &&	0.960 \\
    Ours-hybrid && 0.983 &	0.938 &	0.975 &	0.982 &	0.982 &	0.950 &	0.988 &	0.897 &&	0.962 \\
    % Ours-explicit && 0.976 &	0.953 &	0.977 &	0.981 &	0.980 &	0.968 &	0.985 &	0.892 &&	0.964 \\
    % Ours-hybrid   && 0.980 &	0.953 &	0.980 &	0.982 &	0.983 &	0.971 &	0.987 &	0.899 &&	0.967 \\
    INGP          && -     & -     & -     & -     & -     & -     & -     & -     && -     \\
    TensoRF       && 0.985 & 0.937 & 0.982 & 0.982 & 0.983 & 0.952 & 0.988 & 0.895 && 0.963 \\
    Plenoxels     && 0.977 & 0.933 & 0.976 & 0.980 & 0.975 & 0.949 & 0.985 & 0.890 && 0.958 \\
    JAXNeRF       && 0.975 & 0.929 & 0.970 & 0.978 & 0.970 & 0.955 & 0.983 & 0.868 && 0.954 \\
    \bottomrule
  \end{tabular}
  \caption{\textbf{Full results on static synthetic scenes~\cite{nerf}.} Dashes denote values that were not reported in prior work.}
  \label{tab:fullsynthetic}
\end{table*}

\renewcommand{\tabcolsep}{6pt}
\begin{table*}
  \centering
  \begin{tabular}{llcccccccclc}
    \multicolumn{12}{c}{PSNR $\uparrow$} \\
    \toprule
    && Room & Fern & Leaves & Fortress & Orchids & Flower & T-Rex & Horns && Mean \\ 
    \cmidrule(){1-1} \cmidrule(){3-10} \cmidrule(){12-12}
    Ours-explicit              && 32.72 & 24.87 & 21.07 & 31.34 & 19.89 & 28.37 & 27.54 & 28.40 && 26.78 \\
    Ours-hybrid                && 32.64 & 25.38 & 21.30 & 30.44 & 20.26 & 28.67 & 28.01 & 28.64 && 26.92 \\
    NeRF~\cite{nerf}           && 32.70 & 25.17 & 20.92 & 31.16 & 20.36 & 27.40 & 26.80 & 27.45 && 26.50 \\
    Plenoxels~\cite{plenoxels} && 30.22 & 25.46 & 21.41 & 31.09 & 20.24 & 27.83 & 26.48 & 27.58 && 26.29 \\
    TensoRF (L)~\cite{tensorf} && 32.35 & 25.27 & 21.30 & 31.36 & 19.87 & 28.60 & 26.97 & 28.14 && 26.73 \\
    DVGOv2~\cite{dvgo}         && -     & -     & -     & -     & -     & -     & -     & -     && 26.34 \\
    \bottomrule
    
    \multicolumn{12}{c}{SSIM $\uparrow$} \\
    \toprule
    && Room & Fern & Leaves & Fortress & Orchids & Flower & T-Rex & Horns && Mean \\ 
    \cmidrule(){1-1} \cmidrule(){3-10} \cmidrule(){12-12}
    Ours-explicit              && 0.955 & 0.809 & 0.738 & 0.898 & 0.665 & 0.867 & 0.909 & 0.884 && 0.841 \\
    Ours-hybrid                && 0.957 & 0.828 & 0.746 & 0.890 & 0.676 & 0.872 & 0.915 & 0.892 && 0.847 \\
    NeRF~\cite{nerf}           && 0.948 & 0.792 & 0.690 & 0.881 & 0.641 & 0.827 & 0.880 & 0.828 && 0.811 \\
    Plenoxels~\cite{plenoxels} && 0.937 & 0.832 & 0.760 & 0.885 & 0.687 & 0.862 & 0.890 & 0.857 && 0.839 \\
    TensoRF (L)~\cite{tensorf} && 0.952 & 0.814 & 0.752 & 0.897 & 0.649 & 0.871 & 0.900 & 0.877 && 0.839 \\
    DVGOv2~\cite{dvgo}         && -     & -     & -     & -     & -     & -     & -     & -     && 0.838 \\
    \bottomrule
  \end{tabular}
  \caption{\textbf{Full results on static forward-facing scenes \cite{llff}.} Dashes denote values that were not reported in prior work.}
  \label{tab:full_llff}
\end{table*}

\renewcommand{\tabcolsep}{6pt}
\begin{table*}
  \centering
  \begin{tabular}{llcccccccclc}
    \multicolumn{12}{c}{PSNR $\uparrow$} \\
    \toprule
     & & Hell Warrior & Mutant & Hook & Balls  & Lego & T-Rex & Stand Up & Jumping Jacks & & Mean \\ 
     \cmidrule(){1-1} \cmidrule(){3-10} \cmidrule(){12-12}
     % These use half resolution (to match prior work), and time smoothness 0.01
     Ours-explicit && 25.60 &	33.56 &	28.21 &	38.99 &	25.46 &	31.28 &	33.27 &	32.00 &&	31.05 \\
     Ours-hybrid && 25.70 &	33.79 &	28.50 &	41.22 &	25.48 &	31.79 &	33.72 &	32.64 &&	31.61 \\
     % These use full resolution and time smoothness 0.1
     % Ours-explicit && 24.77 &	32.43 &	27.84 &	38.59 &	25.25 &	30.45 &	32.50 &	31.26	&& 30.39 \\
     % Ours-hybrid && 24.81 &	32.59 &	28.13 &	40.33 &	25.27 &	30.75 &	33.17 &	31.64 &&	30.84 \\
     D-NeRF \cite{dnerf} && 25.02 & 31.29	& 29.25 & 32.80	& 21.64	& 31.75 &	32.79 &	32.80 &&	29.67 \\
    T-NeRF \cite{dnerf} &&	23.19 &	30.56 &	27.21 &	32.01	& 23.82 &	30.19 &	31.24 &	32.01 &&	28.78 \\
    % NeRF \cite{nerf} && 13.52 &	20.31 &	16.65 &	18.28	& 20.3 &	24.49 &	18.19 &	18.28 &&	18.75 \\
Tensor4D \cite{tensor4d} && - & - & - & - &	26.71 &	- & 36.32	& 34.43 &&	- \\
TiNeuVox \cite{tineuvox} && 28.17	& 33.61 &	31.45 &	40.73 &	25.02 &	32.70 &	35.43 &	34.23 &&	32.67 \\
V4D \cite{v4d}	&& 27.03	& 36.27 &	31.04 &	42.67 &	25.62 &	34.53 &	37.20 &	35.36 &&	33.72 \\
    \bottomrule
    \multicolumn{12}{c}{SSIM $\uparrow$} \\
    \toprule
     & & Hell Warrior & Mutant & Hook & Balls  & Lego & T-Rex & Stand Up & Jumping Jacks & & Mean \\ 
     \cmidrule(){1-1} \cmidrule(){3-10} \cmidrule(){12-12}
     % These use half resolution (to match prior work), and time smoothness 0.01
     Ours-explicit && 0.951 &	0.982 &	0.951 &	0.989 &	0.947 &	0.980 &	0.980 &	0.974 &&	0.969 \\
     Ours-hybrid && 0.952 &	0.983 &	0.954 &	0.992 &	0.948 &	0.981 &	0.983 &	0.977 &&	0.971 \\
     % These use full resolution and time smoothness 0.1
     % Ours-explicit && 0.948 &	0.968 &	0.941 &	0.986 &	0.934 &	0.972 &	0.973 &	0.966 &&	0.961 \\
     % Ours-hybrid && 0.950 &	0.968 &	0.945 &	0.989 &	0.936 &	0.973 &	0.977 &	0.969 &&	0.963 \\
     D-NeRF \cite{dnerf} && 0.95 &	0.97 &	0.96 &	0.98 &	0.83 &	0.97 &	0.98 &	0.98 &&	0.95 \\
    T-NeRF \cite{dnerf} &&	0.93 &	0.96 &	0.94 &	0.97 &	0.90 &	0.96 &	0.97 &	0.97 &&	0.95 \\
    % NeRF \cite{nerf} && 13.52 &	20.31 &	16.65 &	18.28	& 20.3 &	24.49 &	18.19 &	18.28 &&	18.75 \\
Tensor4D \cite{tensor4d} && - & - & - & - &	0.953 &	- & 0.983 & 0.982 &&	- \\
TiNeuVox \cite{tineuvox} && 0.97 &	0.98 &	0.97 &	0.99 &	0.92 &	0.98 &	0.99 &	0.98 &&	0.97 \\
V4D \cite{v4d}	&& 0.96 &	0.99 &	0.97 &	0.99 &	0.95	& 0.99	& 0.99	& 0.99	&& 0.98 \\
    \bottomrule
  \end{tabular}
      \vspace{-0.1cm}
    \begin{flushleft}
  % {\footnotesize $^\dagger$ Very recent/concurrent work. Tensor4D was released in November 2022, TiNeuVox in September 2022, and V4D in October 2022. }
  \end{flushleft}
  \vspace{-0.4cm}
  \caption{\textbf{Full results on monocular ``teleporting-camera'' dynamic scenes.} We use the synthetic scenes from D-NeRF \cite{dnerf}, which we refer to as monocular ``teleporting-camera'' because although there is a single training view per timestep, the camera can move arbitrarily between adjacent timesteps. Dashes denote unreported values. TiNeuVox trains in 30 minutes, V4D in 4.9 hours, D-NeRF in 2 days, and Tensor4D for an unspecified duration (Tensor4D reports iterations rather than time). Our reported results were obtained after roughly 1 hour of optimization on a single GPU. Like D-NeRF and TiNeuVox, we train and evaluate using half-resolution images (400 by 400 pixels).}
  \label{tab:fulldnerf}
\end{table*}

\renewcommand{\tabcolsep}{4pt}
\begin{table*}
  \centering
  \begin{tabular}{llcccccclc}
    \multicolumn{10}{c}{PSNR $\uparrow$} \\
    \toprule
    && Coffee Martini & Spinach & Cut Beef & Flame Salmon\footnotemark[1] & Flame Steak & Sear Steak && Mean \\ 
    \cmidrule(){1-1} \cmidrule(){3-8} \cmidrule(){10-10}
    Ours-explicit                && 28.74 & 32.19 & 31.93 & 28.71 & 31.80 & 31.89 && 30.88 \\
    Ours-hybrid                  && 29.99 & 32.60 & 31.82 & 30.44 & 32.38 & 32.52 && 31.63 \\
    LLFF~\cite{llff}                         && -     & -     & -     & 23.24 & -     & -     && -     \\
    DyNeRF~\cite{dynerf}         && -     & -     & -     & 29.58 & -     & -     && -     \\
    MixVoxels-L$^\dagger$~\cite{mixvoxels} && 29.36 & 31.61 & 31.30 & 29.92 & 31.21 & 31.43 && 30.80 \\
    \bottomrule
  
    \multicolumn{10}{c}{SSIM $\uparrow$} \\
    \toprule
    && Coffee Martini & Cook Spinach & Cut Beef & Flame Salmon\footnotemark[1] & Flame Steak & Sear Steak && Mean \\ 
    \cmidrule(){1-1} \cmidrule(){3-8} \cmidrule(){10-10}
    Ours-explicit   && 0.943 & 0.968 & 0.965 & 0.942 & 0.970 & 0.971 && 0.960 \\
    Ours-hybrid     && 0.953 & 0.966 & 0.966 & 0.953 & 0.970 & 0.974 && 0.964 \\
    LLFF            && -     & -     & -     & 0.848 & -     & -     && -     \\
    DyNeRF          && -     & -     & -     & 0.961 & -     & -     && -     \\
    MixVoxels-L     && 0.946 & 0.965 & 0.965 & 0.945 & 0.970 & 0.971 && 0.960 \\
    \bottomrule
  \end{tabular}
    \vspace{-0.1cm}
    \begin{flushleft}
  {\footnotesize $^\dagger$ Very recent/concurrent work. MixVoxels was released in December 2022.}
  {\footnotesize \footnotemark[1]Using the first 10 seconds of the 30 second long video.}
  \end{flushleft}
  \vspace{-0.4cm}
  \caption{\textbf{Full results on multiview dynamic scenes~\cite{dynerf}.} Dashes denote unreported values. Note that our method optimizes in less than 4 GPU hours, whereas DyNeRF trains on 8 GPUs for a week, approximately 1344 GPU hours.}
  \label{tab:fullvideo}
\end{table*}

\renewcommand{\tabcolsep}{6pt}
\begin{table*}
  \centering
  \begin{tabular}{llccclc}
    \multicolumn{7}{c}{PSNR $\uparrow$} \\
    \toprule
     & & Brandenburg Gate & Sacre Coeur & Trevi Fountain & & Mean \\ 
     \cmidrule(){1-1} \cmidrule(){3-5} \cmidrule(){7-7}
     Ours-explicit && 24.85 & 19.90 & 22.00 && 22.25 \\
     Ours-hybrid && 25.49 & 20.61 & 22.67 && 22.92 \\
     NeRF-W~\cite{martinbrualla2020nerfw} && 29.08 & 25.34 & 26.58 && 27.00 \\
     NeRF-W (public)$^\dagger$ && 21.32 & 19.17 & 18.61 && 19.70 \\
     LearnIt \cite{learnit} && 19.11 & 19.33 & 19.35 && 19.26 \\
    \bottomrule 
    \multicolumn{7}{c}{MS-SSIM $\uparrow$} \\
    \toprule
    & & Brandenburg Gate & Sacre Coeur & Trevi Fountain & & Mean \\ 
    \cmidrule(){1-1} \cmidrule(){3-5} \cmidrule(){7-7}
    Ours-explicit && 0.912 & 0.821 & 0.845 && 0.859 \\
    Ours-hybrid && 0.924 & 0.852 & 0.856 && 0.877 \\
    NeRF-W && 0.962 & 0.939 & 0.934 && 0.945 \\
    Nerf-W (public)$^\dagger$ && 0.845 & 0.752 & 0.694 && 0.764 \\
    LearnIt && - & - & - && - \\ \bottomrule

  \end{tabular}
    \vspace{-0.1cm}
    \begin{flushleft}
  {\footnotesize $^\dagger$ Open-source version \url{https://github.com/kwea123/nerf_pl/tree/nerfw} where we implement the test-time optimization ourselves exactly as for \modelname{}. NeRF-W code is not public.}
  \end{flushleft}
  \vspace{-0.4cm}
  \caption{\textbf{Full results on phototourism scenes.} Note that our results were obtained after about 35 GPU minutes, whereas NeRF-W trains with 8 GPUs for two days, approximately 384 GPU hours. }
  \label{tab:fullphototourism}
\end{table*}

\end{document}